\documentclass[sigconf, anonymous=false]{acmart}

\AtBeginDocument{%
  \providecommand\BibTeX{{%
    \normalfont B\kern-0.5em{\scshape i\kern-0.25em b}\kern-0.8em\TeX}}}

\setcopyright{acmcopyright}
\copyrightyear{2024}
\acmYear{2024}
\setcopyright{acmlicensed}

\acmConference[WSDM '24] {Proceedings of the 17th ACM International Conference on Web Search and Data Mining}{March 4--8, 2024}{Merida, Mexico.}

\acmBooktitle{Proceedings of the 17th ACM International Conference on Web Search and Data Mining (WSDM '24), March 4--8, 2024, Merida, Mexico} 
\acmPrice{15.00}
\acmISBN{979-8-4007-0371-3/24/03}
\acmDOI{10.1145/3616855.3635843}

\usepackage{algorithm, algorithmic}
\usepackage{multirow}

\usepackage{amssymb}
\usepackage{subfigure}
\usepackage{adjustbox}
\usepackage{bbding}
\usepackage{balance}

\begin{document}

\settopmatter{printacmref=true} 

\newcommand{\ourmethod}[1]{}
\renewcommand{\ourmethod}[1]{\texttt{LMBot}}

\title{\ourmethod{}: Distilling Graph Knowledge into Language Model for Graph-less Deployment in Twitter Bot Detection}

\author{Zijian Cai}
\email{2205114706@stu.xjtu.edu.cn}
\affiliation{
  \institution{Xi’an Jiaotong University}
  \city{Xi'an}
  \state{Shaanxi}
  \country{China}
}

\author{Zhaoxuan Tan}
\email{ztan3@nd.edu}
\affiliation{
  \institution{University of Notre Dame}
  \city{Notre Dame}
  \state{IN}
  \country{USA}
}

\author{Zhenyu Lei}
\email{vjd5zr@virginia.edu}
\affiliation{
  \institution{University of Virginia}
  \city{Charlottesville}
  \state{VA}
  \country{USA}
}

\author{Zifeng Zhu}
\email{zivenzhu@stu.xjtu.edu.cn}
\affiliation{
  \institution{Xi’an Jiaotong University}
  \city{Xi'an}
  \state{Shaanxi}
  \country{China}
}

\author{Hongrui Wang}
\email{wanghongrui@stu.xjtu.edu.cn}
\affiliation{
  \institution{Xi’an Jiaotong University}
  \city{Xi'an}
  \state{Shaanxi}
  \country{China}
}

\author{Qinghua Zheng}
\email{qhzheng@mail.xjtu.edu.cn}
\affiliation{
  \institution{Xi’an Jiaotong University}
  \city{Xi'an}
  \state{Shaanxi}
  \country{China}
}

\author{Minnan Luo}
\authornote{Corresponding author: Minnan Luo, School of Computer Science and Technology, Xi’an Jiaotong University, Xi’an 710049, China.}
\email{	minnluo@xjtu.edu.cn}
\affiliation{
  \institution{Xi’an Jiaotong University}
  \city{Xi'an}
  \state{Shaanxi}
  \country{China}
}

\renewcommand{\shortauthors}{Zijian Cai et al.}

\begin{abstract}
 As malicious actors employ increasingly advanced and widespread bots to disseminate misinformation and manipulate public opinion, the detection of Twitter bots has become a crucial task. 
 Though graph-based Twitter bot detection methods achieve state-of-the-art performance, we find that their inference depends on the neighbor users multi-hop away from the targets, and fetching neighbors is time-consuming and may introduce sampling bias. At the same time, our experiments reveal that after finetuning on Twitter bot detection task, pretrained language models achieve competitive performance while do not require a graph structure during deployment. Inspired by this finding, we propose a novel bot detection framework \ourmethod{}\footnote{The code is available at \url{https://github.com/czjdsg/LMBot}} that distills the graph knowledge into language models (LMs) for graph-less deployment in Twitter bot detection to combat data dependency challenge. Moreover, \ourmethod{} is compatible with graph-based and graph-less datasets. Specifically, we first represent each user as a textual sequence and feed them into the LM for domain adaptation. For graph-based datasets, the output of LM serves as input features for the GNN, enabling \ourmethod{} to optimize for bot detection and distill knowledge back to the LM in an iterative, mutually enhancing process. Armed with the LM, we can perform graph-less inference with graph knowledge, which resolves the graph data dependency and sampling bias issues. For datasets without graph structure, we simply replace the GNN with an MLP, which also shows strong performance.
 Our experiments demonstrate that \ourmethod{} achieves state-of-the-art performance on four Twitter bot detection benchmarks. Extensive studies also show that \ourmethod{} is more robust, versatile, and efficient compared to existing graph-based Twitter bot detection methods.
\end{abstract}




\maketitle
\begin{figure}[t]
    \centering
    \includegraphics[width=\linewidth]{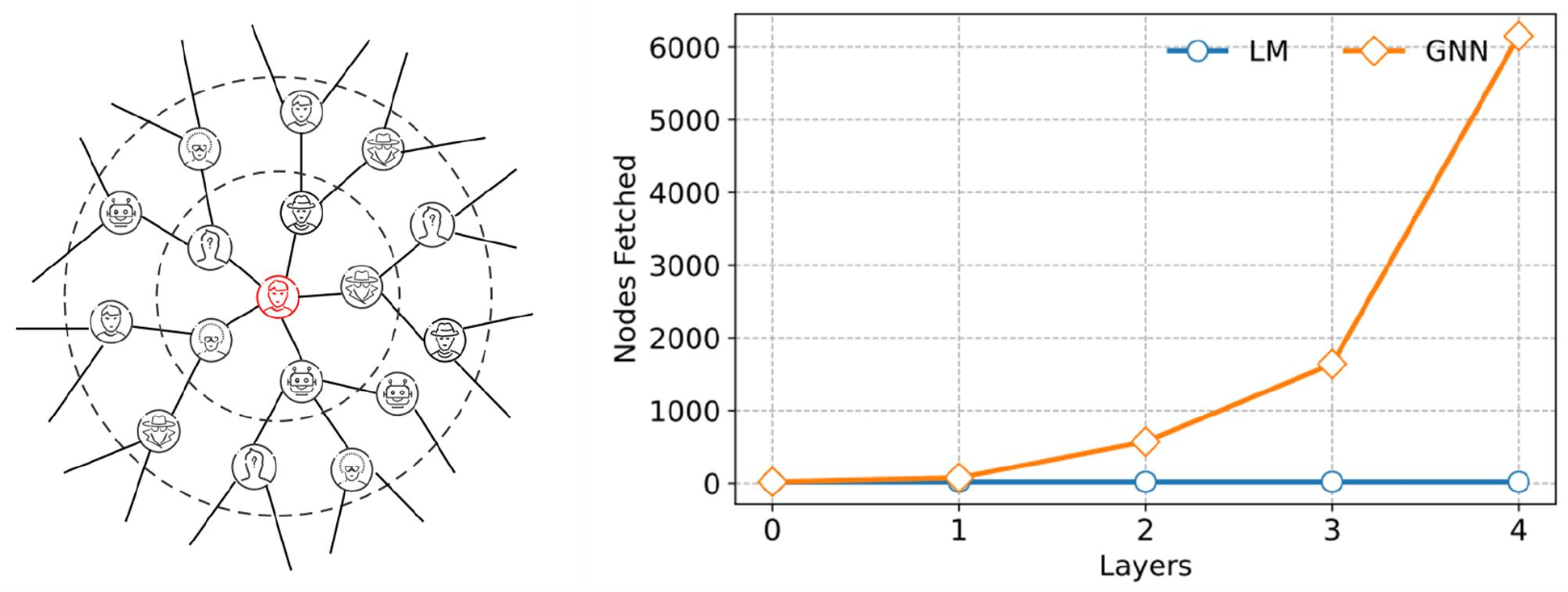}
    
    \caption{A typical Twittersphere in TwiBot-20 dataset (left) and the number of fetched users when doing inference for 20 users using language model and graph neural networks with the increasing number of layers (right).}
    \label{fig:teaser}
\end{figure}

\section{Introduction}
Twitter bots are automated accounts that perform various functionalities. They are manipulated by malignant actors and pose significant risks, particularly in terms of disseminating false information and facilitating fraudulent activities. They have been continuously involved in spreading misinformation \cite{howard2017junk}, manipulating opinion \cite{ferrara2016rise}, and invading privacy by inferring sensitive details from tweets \cite{volkova2015inferring}. The increasing prevalence and advancing techniques could greatly amplify their influence if left unregulated. Therefore, it is of great significance and urgency to develop effective Twitter bot detection methods.

Researchers have devoted significant efforts to detecting Twitter bots, and the existing methods can be categorized into three types: feature-based, text-based, and graph-based. Early methods that focus on designing efficient features are primarily feature-based and text-based methods. Feature-based methods extract features from users' metadata and tweets, which are then fed into random forests for bot identification \cite{echeverri2018lobo, beskow2018bot, kouvela2020bot}. However, bot manipulators began to be aware of the selected features and bots can disguise themselves from feature-based methods by tampering with bot metadata. Text-based methods apply NLP techniques to encode users' descriptions and tweet information for bot detection \cite{hayawi2022deeprobot, wu2023bottrinet}. However, text-based methods can be easily deceived when bots steal tweets posted by genuine users \cite{liu2023botmoe}. Later, researchers proposed graph-based methods that leverage the intrinsic graph structure in Twittersphere and apply network science and geometric deep learning techniques for bot detection \cite{feng2021botrgcn, feng2022heterogeneity}. The state-of-the-art performance and extensive studies on graph-based methods indicate the necessity of graph structure for Twitter bot detection. 
Despite the impressive accuracy achieved by graph-based methods, we argue that making inferences with graph structure is time-consuming. That being said, for each target user, we need to fetch its multi-hop neighbor users' information. Given Twitter API's slow response rate and strict rate limits, querying neighborhood information is expensive and burdens the real-world deployment of Twitter bot detection applications. As is shown in Figure \ref{fig:teaser}, in the TwiBot-20 dataset \cite{feng2021twibot}, with the increase of Graph Neural Network (GNN) layers, there is a concurrent rise in hops of neighboring users, leading to an exponential growth in the number of neighbors to be queried.
Moreover, restricted by the Twitter API, researchers are only able to obtain 20 neighboring users through random sampling per query, which also introduces the graph sampling bias problem \cite{leskovec2006sampling, zeng2019graphsaint}. Hence, developing a Twitter bot detection framework that can make inferences without querying graph structure is essential for efficient and reliable real-world applications.



\begin{figure}[t]
    \centering\includegraphics[width=0.7\linewidth]{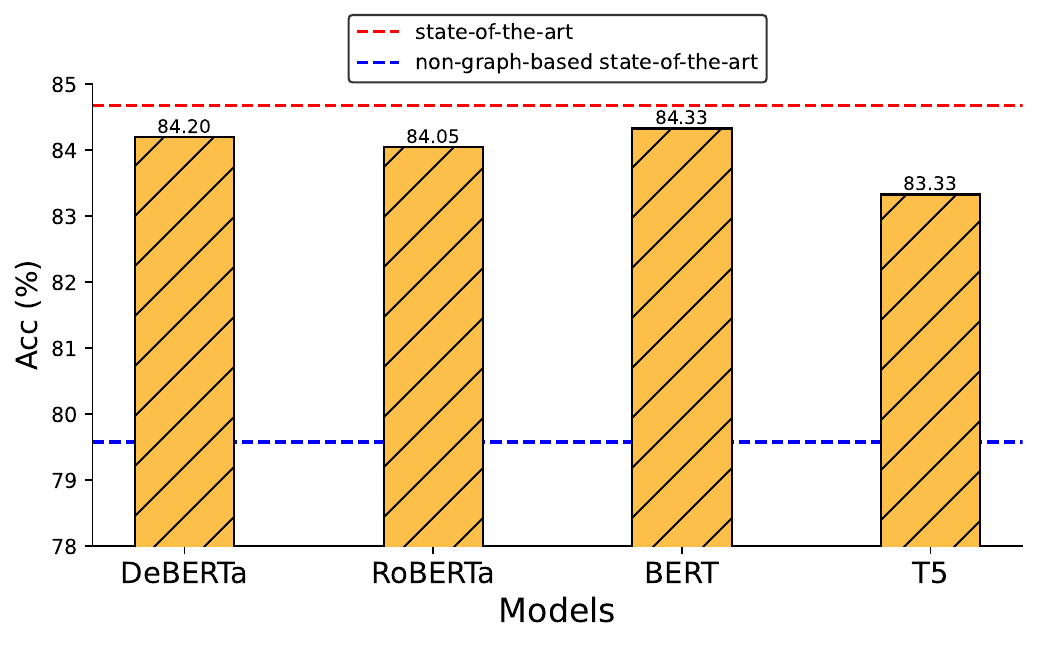}
    \caption{The performance of finetuned language models on TwiBot-20 dataset. Simply finetuning language models not only outperforms the state-of-the-art non-graph-based methods, but also closely rivals the state-of-the-art performance.}
    \label{fig:finetune}
\end{figure}

In our pursuit of creating a Twitter bot detection framework capable of conducting inference without relying on graph structure, we first scrutinize existing non-graph-based methods and find that text-based methods perform the best \cite{feng2022twibot}. However, we notice that these methods mostly process user text features using frozen pretrained language models, which greatly limits the expressive power of language models and their adaptation to the Twitter bot detection domain.
To address this problem, we first finetune the language models for the Twitter bot detection task. Our findings are quite remarkable – the finetuned models exhibited impressive performance. As is demonstrated in Figure \ref{fig:finetune}, simply finetuning language models can achieve competitive performance that already outperforms the state-of-the-art non-graph-based methods and is close to the state-of-the-art performance achieved by graph-based methods. These results indicate that the ability of language models for bot detection is largely underestimated. Moreover, these LMs do not require neighboring (graph) information for inference, as is shown in Figure \ref{fig:teaser}, they emerge as a promising proxy for implementing efficient and effective graph-less Twitter bot detection frameworks.


However, merely finetuning language models is yet to be perfect, as there remains a gap between the performance of vanilla finetuned language models and state-of-the-art methods. Given the state-of-the-art performance of graph-based methods, and previous studies indicating that the graph structure in Twittersphere is highly effective in tackling various challenges, such as bot communities \cite{liu2023botmoe}, and bot disguise \cite{feng2021botrgcn, ye2023hofa}. As a result, we are convinced that the graph structure in Twittersphere is essential for Twitter bot detection. Nevertheless, the question of how to incorporate graph information into language models for bot detection remains underexplored.



To efficiently incorporate graph information into language models for Twitter bot detection, we propose \ourmethod{}, a novel framework that iteratively distills knowledge from graph neural networks into language models.
Specifically, we first extract each user's metadata, descriptions, and tweets into a textual sequence for unified user representation. We then feed these text sequences into the LM and conduct finetuning for domain adaption, allowing the LM to acquire domain knowledge of the Twittersphere. Next, we iteratively distill knowledge and inductive bias from GNN into LM. We first update GNN's parameters, where the output of LM is leveraged as the initial feature for GNN, providing GNN with better input \emph{w.r.t} Twitter bot detection task. After updating GNN's parameters, we then update LM's learnable parameters by using GNN's prediction logits to guide LM's training through knowledge distillation, therefore incorporating graph information into LM. \ourmethod{} performs these two steps iteratively, allowing the GNN and LM to mutually enhance and iteratively distill graph information into the language model. In the inference stage, either LM or GNN can be used for Twitter bot detection. Armed with LM, we don't need to fetch neighboring information for deployment, therefore saving lots of time waiting for API response and avoiding graph sampling bias. For instance, in a deployment scenario as the one illustrated in Figure \ref{fig:teaser}, inferring information about 20 new users takes merely about 60 milliseconds with the proposed \ourmethod{}. In contrast, using a GNN would have to fetch data for 552 additional neighbors (2-hop neighbors), which could take approximately 400 seconds using the Twitter API. For datasets without graph structure,  we are inspired by the success of feature-based methods that only leverage MLP \cite{hayawi2022deeprobot} for bot identification. By simply replacing GNN with MLP, we find \ourmethod{} framework continues to deliver state-of-the-art performance.

Our experiments demonstrate that \ourmethod{} achieves state-of-the-art performance on four widely-acknowledged Twitter bot detection benchmarks, outperforming 10 competitive Twitter bot detection baselines. In addition, we study different combinations of LMs and GNNs, and find that the selection of LMs is more robust, while GNNs are sensitive, with heterogeneous GNNs outperforming their homogeneous counterparts. We also analyze the convergence process of the LM and GNN, validate the mutually enhancing process between them, and confirm the necessity of domain adaptation for the language model's fast convergence speed and stability. Moreover, We study the prediction consistency between LM and GNN, confirming that LM can give similar predictions as GNN even without graph structure in the inference stage. This result demonstrates that LM can effectively learn the inductive bias in GNN. Further studies also reveal the high data efficiency of \ourmethod{}, where \ourmethod{} consistently achieves high performance and exhibits lower volatility when the ground truth label and edge in the Twitter network decrease.

\begin{figure*}[t]
    \centering
    \includegraphics[width=0.8\linewidth]{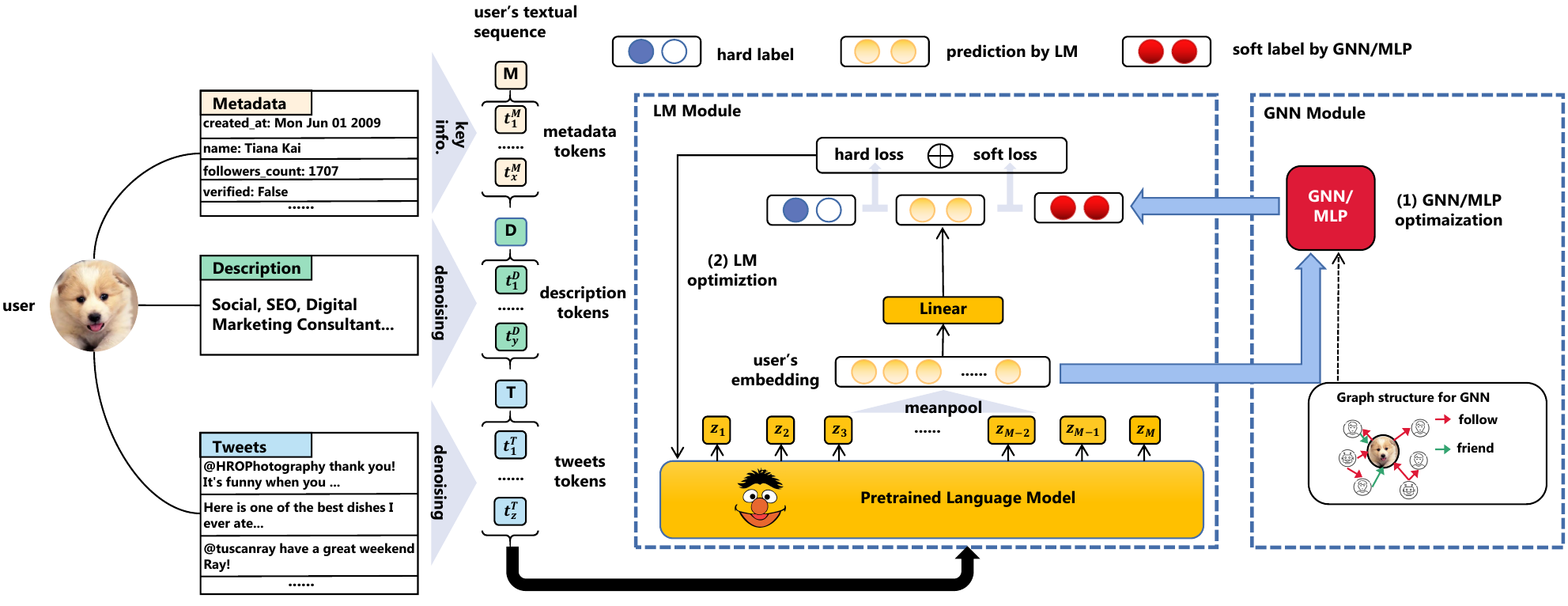}
    \caption{Overview of our proposed framework \ourmethod{}. \ourmethod{} first represents the user as a textual sequence, serving as the input of LM. \ourmethod{} then iteratively distills graph structure knowledge in GNN into LM, where LM provides GNN with user embeddings and GNN provides LM with soft labels.}
    \label{fig:overview}
\end{figure*}
Our main contributions are summarized as follows:
\begin{itemize}
    \item We are the first to discover that language model-based Twitter bot detection methods are significantly underestimated. Specifically, we finetune language models for the Twitter bot detection task and find them to outperform all non-graph-based methods and rival the performance of the state-of-the-art graph-based methods.
    \item We propose \ourmethod{}, the first graph-less bot detection framework that incorporates graph knowledge in LM by iteratively distilling graph information extracted by GNN into LM. Armed with \ourmethod{}, neither neighboring information nor graph structure is required during inference, resolving the challenge of data dependency and sampling bias in graph-based Twitter bot detection deployment.
    \item Our experiments show that \ourmethod{} achieves state-of-the-art on four widely used Twitter bot benchmarks, even without graph structure. 
    Extensive studies further confirm the effectiveness, efficiency, and robustness of \ourmethod{}.
\end{itemize}

\section{Related Work}  
\subsection{Twitter Bot Detection}
Twitter bot detection refers to detecting and identifying bot accounts on the Twitter platform. Bot accounts are usually automatically controlled by the software and imitate genuine users to post tweets. Identifying these bot accounts can preserve the integrity of online discourse, and combat misinformation and conspiracy theories. Existing methods are mostly feature-based, text-based, and graph-based \cite{feng2022twibot}. 

\textbf{Feature-Based Methods.} 
These methods extract features from the user's metadata and tweets and leverage a classifier for bot detection. \citet{echeverri2018lobo} extract 10 features from users’ metadata and 19 feature from users' tweets and adopts a random forest for classification. \citet{beskow2018bot} extracts features from user attributes, network attributes, contents, and timing information and exploits random forest as the classifier. \citet{kouvela2020bot} extract 36 features from each user's metadata and the latest 20 tweets and classifies users with random forest. \citet{kudugunta2018deep} extract features from users’ metadata and combines synthetic minority oversampling (SMOTE) with undersampling techniques. \citet{miller2014twitter} cluster human accounts using DBSCAN and K-MEANS to obtain human clusters in training. In testing, users not belonging to any human clusters are considered bots. \citet{abreu2020twitter} choose five essential Twitter user features and classifies users with four machine learning algorithms.

\textbf{Text-Based Methods.} 
These methods leverage NLP techniques to encode the users' textual information for bot detection. \citet{hayawi2022deeprobot} encode users' textual information with LSTM and fuse it with features extracted from metadata using a dense layer. \citet{wu2023bottrinet} produce word, sentence, and account embeddings with a triplet network that refines raw embeddings using metric learning techniques. \citet{guo2021social} construct a heterogeneous graph of the corpus, where words and user descriptions embedded by pretrained BERT are graph nodes. The final prediction is the combination of BERT and GCN output. \citet{wei2019twitter} use pretrained GloVe word vectors on Twitter as word embedding and multiple layers of bidirectional LSTMs for bot detection.

\textbf{Graph-Based Methods.} 
These methods model Twitter as graphs and apply network science and geometric deep learning techniques for Twitter bot detection. \citet{feng2021satar} utilize user semantics, attributes, and neighborhood information, pretraining on abundant self-supervised data and finetuning for different bot detection tasks. \citet{feng2021botrgcn} first apply heterogeneous graph neural networks RGCN \cite{schlichtkrull2018modeling} to the bot detection task, achieving breakthrough performance. \citet{feng2022heterogeneity} propose relational graph transformers to model heterogeneous user influence and use semantic attention networks to aggregate cross-relation messages. \citet{ali2019detect} extract 6 numerical features for each user, and the extracted features are fed into GCN layers for classification. \citet{yang2022rosgas} use reinforcement learning and self-supervision to search GNN architectures, learns user subgraph embeddings from heterogeneous networks for social bot detection. \citet{yang2013empirical} design features from profile, content, graph, neighbor, timing, and automation and classifies with random forest. HOFA \cite{ye2023hofa} discovers the homophily disguise challenge and combat it with graph data augmentation and frequency adaptive attention mechanism.

\textbf{Multi-Modal Methods.} 
Recent studies have used multi-modal methods that fuse multiple information sources to detect Twitter bots. Instead of relying on a single source of information such as user attributes, behavioral features, or tweet content alone, multiple sources of information are fused together for analysis. \citet{ng2022botbuster} use a mixture of experts approach where each expert analyzes a specific type of account information and their results are combined to determine if the account is a bot. \citet{lei2022bic} integrate text and graph modality with a text-graph interaction module, with additional functionality to detect advanced bots with a semantic consistency model. \citet{liu2023botmoe} use multiple user information types in community-aware mixture-of-experts layers to improve the detection of deceptive bots and adapt to different Twitter communities. \citet{tan2023botpercent} leverage the largest ensemble to probe the bot percentage on Twitter.

Upon analyzing existing methods, we found that graph-based methods generally outperform non-graph-based methods, indicating the importance of social network structural information in the Twitter bot detection task. Although graph-based methods achieve state-of-the-art performance, they struggle to be deployed in real-world social networks because they require fetching exponential neighboring information when detecting target users, which is heavily time-consuming. In this work, we proposed \ourmethod{} to leverage language models as a proxy for graph-less Twitter bot detection inference and integrate Twitter's graph context information in the training stage by iteratively distilling the graph knowledge into LM.  This integration allows the LM component of \ourmethod{} to efficiently conduct Twitter bot detection without fetching neighboring data, thereby offering a viable and effective solution for practical deployment in real-world scenarios.

\subsection{Knowledge Distillation}

Knowledge Distillation (KD) \cite{hinton2015distilling} is a model compression technique where a small student model learns from a large trained teacher model. The student model mimics the teacher to achieve similar performance but with fewer parameters and computations. Among the existing knowledge distillation methods, there are distillations between the same type of models and different types of models 

\textbf{Distillations between the same type of models.} 
This is conducted in order to reduce the number of parameters. For example, \citet{jiao2019tinybert} propose a two-stage framework that performs distillation during pretraining and task training, enabling TinyBERT to capture BERT's general and task knowledge.  \citet{sanh2019distilbert} introduce a triple loss combining language modeling, distillation, and cosine-distance losses to distill BERT to a smaller general-purpose language model DistilBERT during pretraining. \citet{yan2020tinygnn} propose a small GNN with explicit peer neighborhood modeling and implicit knowledge distillation from a deeper GNN to obtain high-performance and fast node representations. 

\textbf{Distillations between different types of models.} 
Naturally, knowledge distillation can also be performed between different types of models. When there are differences in the structure or mechanisms of the two models, the knowledge background and global understanding provided by the teacher model are difficult for the student model to acquire on its own, so this type of knowledge distillation is also significant. For example, GNN can be distilled into MLP. \citet{zhang2021graph} propose graph-less neural networks (GLNNs) by distilling GNN knowledge into MLPs, which have no graph dependency and infer faster, to address GNN scalability while maintaining accuracy.  \citet{guo2022linkless} propose a relational KD framework, Linkless Link Prediction (LLP) that distills relational knowledge which is centered around each  node to the student MLP. \citet{zhao2022learning} propose an iterative framework that can alternately train LM and GNN through pseudo-labels, where they mutually boost each other's performance.

Despite much research progress has been achieved in GNN distillation, distilling text-rich heterogeneous information network information into language models remains under-explored. In this work, we treat LM as the student model and GNN as the teacher model and distill the graph structure knowledge in GNN into LM. At the same time, LM provides embeddings for GNN. After iteratively learning the knowledge from GNN, a graph-free LM can achieve higher performance compared to graph-based Twitter bot detection methods.


\section{Methodology}
Figure \ref{fig:overview} presents an overview of our proposed framework \ourmethod{}. Specifically, \ourmethod{} first presents user as a textual sequence, which is fed into LM to obtain user embedding. We first conduct LM domain adaptation finetuning on the target task. \ourmethod{} further iteratively distills graph structure knowledge in GNN into LM, where LM provides GNN with user embeddings and GNN provides LM with soft labels. Enhanced by graph structure knowledge, LM shows strong performance and even outperforms graph-based methods.
\subsection{User as a Textual Sequence}
For every Twitter account, we have access to its metadata, tweets, and description. To obtain a universal representation for each user that is compatible with language models' input, we propose a new raw representation method that encodes users as textual sequences, without relying on any complex feature engineering. Specifically, for users' metadata, \ourmethod{} directly extracts the corresponding information from the raw text, such as username, location, and number of followers, then organizes it into a sequence \{metadata\} as follows:
\begin{equation*}
    \rm [M]\{ metadata_1 \}[SEP] \{ metadata_2 \}[SEP]\ldots,
\end{equation*}
where [M] is a special token denoting the beginning of metadata, and [SEP] denotes separation. We apply a similar processing procedure to tweets and descriptions, obtaining a sequence that represents a user:
\begin{equation*}
    \rm [M]\{metadata\}[D]\{description\}[T]\{tweet\},
\end{equation*}
where [M], [D], and [T] are other special tokens to be added to the tokenizer of the corresponding language model. 

In the original tweets and descriptions, there are many noises (\emph{e.g.}, hashtags, mentions, and URLs) that may harm the quality of representation LM generates. Inspired by\citet{nguyen2020bertweet}, we replace all hashtags, mentions, and URLs with \#HASHTAG, @USER, and HTTPURL tokens, and use the \verb|TweetTokenizer|\footnote{\url{https://www.nltk.org/api/nltk.tokenize.casual.html}} to further denoise users' descriptions and tweets. The above operations minimize the noise in the users' textual information, therefore conducive to user representation learning. 
By encoding diverse user information into a textual sequence, we have a universal user representation containing all available information for Twitter bot detection.

\subsection{Iterative Distillation}
To integrate the domain knowledge in Twitter bot detection, we first apply LM domain adaptation finetuning to pretrained language models with user sequences and corresponding labels. Subsequently, \ourmethod{} jointly train LM and GNN/MLP, which could alternately insert graph structure knowledge into the language model. In this framework, LM and GNN/MLP do not update parameters simultaneously. Instead, they update iteratively in LM-step and GNN/MLP-step. Furthermore, we combine these two steps organically using the method of knowledge distillation. 

\subsubsection{LM domain adaptation finetuning}
Considering the difference between the pretraining corpus of language models and the Twitter corpus, we first conduct domain adaptation fine-tuning of the language model on the target task. This step enhances the model stability and accelerates the convergence speed. Firstly, we use LM to encode the user's textual sequence, which can be formulated as:
\begin{equation} \label{eq:0}
    \boldsymbol{z}_i = \frac{1}{M_i}\sum_{j=1} ^{M_i} \mathrm{LM}(\boldsymbol{t}_i)_j,
\end{equation}
where $\boldsymbol{t}_i$ is the textual sequence of $i$-th user, $\mathrm{LM}(\cdot)$ denotes the language model adopted as feature extractor, $M_i$ is the number of tokens in $\boldsymbol{t}_i$. We apply mean pooling to LM output and obtain 768-dimension embeddings for user representations. 

To make predictions, we apply an $L$-layer MLP to reduce the dimension of $\boldsymbol{z}_i$ and project it into a two-way classification space and obtain the predicted logit, which is formulated as:
  \begin{equation}
      \boldsymbol{z}^{(l)}_i = \mathrm{LeakyReLU} (\boldsymbol{W}^{(l)} \cdot \boldsymbol{z}^{(l - 1)}_i+ \boldsymbol{b}^{(l)}),
  \end{equation}
where $\boldsymbol{z}^{(0)}_i$ is the output of LM in Equation \eqref{eq:0}.
We then apply softmax to the output logit $\boldsymbol{z}^{(L)}_i$ to get the prediction $\boldsymbol{\hat{y}}_i$. We then optimize the language model for domain adaptation with the following training objective,
\begin{equation}
        Loss_{\text{LM-FT}} = 
        -\sum_{i \in \mathcal{H}} \sum_{c=1}^C y_{ic} \log (\hat{y}_{ic}) + \lambda_1 \sum _{\theta \in {\Theta_{\text{LM}}}} \theta^2, \label{eq:1}
\end{equation}
where $\mathcal{H}$ is the set of hard labels, $C=2$ is the number of classes in the Twitter bot detection task, $y_{ic} \in \{0, 1\}$ is the ground truth hard label, and $\lambda_1$ is the coefficient of L2 regularization for LM.

\subsubsection{GNN/MLP step}
\ourmethod{} can optionally insert the graph structure knowledge into LM, where we use GNN for datasets with graph structure and MLP for datasets without graph structure.
In the training stage for GNN or MLP updating, GNN/MLP utilizes the embedding $\boldsymbol{z}$ generated by the language model as input and optimizes directly on the Twitter bot detection task. 

When the dataset presents a graph structure, \ourmethod{} takes GNNs to encode graph knowledge, which can be universally expressed as:
\begin{align}
    \boldsymbol{a}_i ^{(l)} = \underset{j \in \mathcal{N}(i)}{\mathrm{AGGERGATE}}&\left[\mathrm{MSG}^{(l)}\left(\boldsymbol{h}_i^{(l-1)}, \boldsymbol{h}_j^{(l-1)}, \boldsymbol{e}_{ji}\right)\right],\\
    \boldsymbol{h}_i ^{(l)} = \mathrm{UPDATE}&\left(\boldsymbol{h}_i^{(l-1)}, \boldsymbol{a}_i^{(l)}\right),
\end{align}
where the $\mathrm{MSG}(\cdot,\cdot,\cdot)$ denotes messages passing from user $i$ to user $j$ through relation, depending on the specific implementation of the GNN layer. $\mathrm{AGGERGATE}(\cdot)$ aggregates messages from all neighbors of node $i$ to get $\boldsymbol{a}_i ^{(l)}$. Finally, $\boldsymbol{h}_i^{(l-1)}$ and $\boldsymbol{a}_i^{(l)}$ are used together to update the representation of node $i$ to get $\boldsymbol{h}_i ^{(l)}$ through $\mathrm{UPDATE}(\cdot,\cdot)$.

After passing $L$ GNN layers, we obtain GNN-based user representations and apply a linear layer to obtain the corresponding user's prediction logit:
\begin{equation}
    \boldsymbol{h}^{o}_i = \boldsymbol{W}_{o} \cdot \mathrm{LeakyReLU}(\boldsymbol{h}^{(L)}_i)+ \boldsymbol{b}_o.
\end{equation}

For datasets without graph structure, we are inspired by the success of the MLP-based method \cite{hayawi2022deeprobot} and replace GNN with an MLP. The MLP module can be expressed as: 
\begin{align}
    \boldsymbol{h}^{(l)}_i &= \boldsymbol{W}^{(l)} \cdot \mathrm{LeakyReLU}(\boldsymbol{h}^{(l-1)}_i)+ \boldsymbol{b}^{(l)},\\
    \boldsymbol{h}^{o}_i &= \boldsymbol{W}_{o} \cdot \mathrm{LeakyReLU}(\boldsymbol{h}^{(L)}_i)+ \boldsymbol{b}_o.
\end{align}
    
The loss function to optimize MLP is similar to Equation \eqref{eq:1}:
\begin{equation} \label{eq:2}
        Loss_{\text{GNN/MLP}} = 
        -\sum_{i \in \mathcal{H}} \sum_{c=1}^C y_{ic} \log (\hat{y}_{ic}) + \lambda_2 \sum _{\theta \in {\Theta_{\text{GNN/MLP}}}} \theta^2.
\end{equation}
After completing GNN/MLP training in this step, we select the GNN/MLP with the best performance on the validation set to generate soft labels for the LM step and distill the knowledge of GNN/MLP into the training of LM in the next step. Soft labels are generated as follows:
\begin{align} \label{eq:3}
    \boldsymbol{\overline{y}}_i = \mathrm{softmax}(\boldsymbol{h}^o_i / T),
\end{align}
where $T$ is the temperature of knowledge distillation, which could control the difficulty of the distillation task \cite{hinton2015distilling}.

\subsubsection{LM step}
In LM training step, GNN acts as the teacher model while LM is the student model to train on the target task. We first get users' embeddings and corresponding classification logits as described in LM domain adaptation finetuning. The training objective in this step can be formulated as:
\begin{equation}\label{eq:4}
\begin{split}
    Loss = & -(1 - \alpha) 
        \sum_{i \in \mathcal{H}} \sum_{c=1}^C y_{ic} \log (\hat{y}_{ic}) \\
    &-\alpha \sum_{i \in \mathcal{S}} \sum_{c=1}^C \overline{y}_{ic} \left[\log (\hat{y}_{ic}) - \log (\overline{y}_{ic})\right]\\
    &+ \lambda_1 \sum _{\theta \in {\Theta_{\text{LM}}}} \theta^2,
\end{split}
\end{equation}
where $\mathcal{S}$ is the soft label set, and $\alpha$ is the coefficient to balance the soft label loss and the hard label loss. Note that the first term is cross entropy loss on hard labels and the second term is Kullback-Leibler divergence  \cite{kullback1951information} loss on soft labels.  

After completion of LM training, we select the LM parameter with the best performance on the validation set to generate embedding for the next step of GNN training.

\subsection{Training and Inference}
The entire training process starts from LM domain adaption, where the training objective is shown in Equation \eqref{eq:1}. After that, \ourmethod{} iteration process begins. Each iteration starts from the GNN/MLP step and updates the parameters of LM and GNN/MLP iteratively. The loss for LM and GNN/MLP training is shown in Equation \eqref{eq:4} and Equation \eqref{eq:2}, respectively.

When the performance of LM or GNN/MLP does not improve on the validation set in a curtain iteration compared with the previous one, we consider that the model has converged and the \ourmethod{} iteration process ends. Further analysis of convergence is presented in Section \ref{sec:1}. After this, these two models can both be used for inference, which are respectively named \ourmethod{}-LM and \ourmethod{}-GNN/MLP. Detailed algorithm is presented in Appendix \ref{app:alg}.

\section{Experiments}
\subsection{Experimental Settings}

\paragraph{Dataset}
We selected 4 widely-acknowledged Twitter bot detection datasets: Cresci-2015 \cite{cresci2015fame}, Cresci-2017 \cite{cresci2017paradigm}, Midterm-2018 \cite{yang2020scalable}, and TwiBot-20 \cite{feng2021twibot}. Cresci-2015 includes account information (i.e., metadata and description) for 5,301 users and their tweets, as well as a graph structure composed of two types of user relationships. The Cresci-2017 dataset contains account information and tweets for 14,368 users, without graph structure. Midterm-2018 is also a graph-less dataset that contains account information for 50,538 users. TwiBot-20 includes account information and tweets for 229,580 users, as well as a graph structure composed of two types of user relationships, but only 11,826 annotated users were actually used. For all datasets, following \cite{shi2023oversampling}, we randomly split users into 1:1:8 ratio for train, valid, and test respectively. We run experiments 5 times and report the average results with standard deviations. Detailed information about the datasets can be found in Appendix \ref{app:dataset} (Table \ref{tab:datasets}).

\paragraph{Baselines}
We compare our \ourmethod{} with the following feature-, text- and graph-based methods: SGBot \cite{yang2020scalable}, BotHunter \cite{beskow2018bot}, LOBO \cite{echeverri2018lobo}, BotRGCN \cite{feng2021botrgcn}, RGT \cite{feng2022heterogeneity}, SimpleHGN \cite{lv2021we}, GLNN \cite{zhang2021graph}, BIC \cite{lei2022bic}, 
BotBuster \cite{ng2022botbuster}. More details about baseline can be found in Appendix \ref{app:baselines}.

\begin{table*}[t]
\renewcommand{\arraystretch}{0.8}
  \centering
  \caption{Average accuracy and F1-score over five runs on four Twitter bot detection datasets. The values in parentheses are standard deviations. The best results are in bold and the second best are underlined. '-' indicates missing information (such as tweets or graph structure) in the dataset to implement the method. $^\ast$ denotes that the results are significantly better than the best baseline method under the student t-test.}
    \begin{tabular}{lcccccccc}
    \toprule[1.5pt]
    \multirow{2}[3]{*}{\textbf{Method}} & \multicolumn{2}{c}{\textbf{Cresci-2015}} & \multicolumn{2}{c}{\textbf{TwiBot-20}} & \multicolumn{2}{c}{\textbf{Cresci-2017}} &
    \multicolumn{2}{c}{\textbf{Midterm-2018}}\\ 
    \cmidrule(r){2-3} \cmidrule(r){4-5} \cmidrule(r){6-7}  \cmidrule(r){8-9}      
& Accuracy   & F1-score   & Accuracy   & F1-score   & Accuracy   & F1-score & Accuracy   & F1-score \\
\midrule[0.75pt]
   \textsc{SGBot} \cite{yang2020scalable}  & 97.76~\small{(0.3)} & 98.23 \small{(0.2)} & 79.58 \small{(0.4)} & 83.51~\small(0.4) & 95.49~\small(0.3) & 97.04~\small(0.2) & 98.61~\small(0.1) & 99.18~\small(0.1) \\
   \textsc{BotHunter} \cite{beskow2018bot} & 96.95~\small(0.6) & 97.58~\small(0.5) & 73.43~\small(0.7) & 78.01~\small(0.9) & 95.76~\small(0.3) & 97.21~\small(0.2) & 98.81~\small(0.0) & 99.29~\small(0.0) \\
   \textsc{Kudugunta} \textit{et al.} \cite{kudugunta2018deep} & 96.11~\small(0.9)& 
   96.88~\small(0.7) &
   58.08~\small(1.7) & 48.07~\small(4.3) &  91.26~\small(0.5) & 93.98~\small(0.4) & 91.86~\small(1.2) & 94.96~\small(0.8) \\
   \textsc{LOBO} \cite{echeverri2018lobo} & 97.76~\small(0.4) & 98.23~\small(0.3) & 76.17~\small(0.5) & 80.64~\small(0.4) & 97.67~\small(0.3) & 98.46~\small(0.2)  & - & - \\
    \textsc{BotRGCN} \cite{feng2021botrgcn} & 98.60~\small(0.1) & 98.90~\small(0.1) & 84.42~\small(0.6) & 86.90~\small(0.1) & - & - & - & - \\
    \textsc{RGT} \cite{feng2022heterogeneity}   & 98.69~\small(0.1) & 98.97~\small(0.1) & 84.70~\small(0.3) & 87.19~\small(0.2) & - & - & - & - \\  
    \textsc{SimpleHGN} \cite{lv2021we} & 98.63~\small(0.1) & 98.92~\small(0.1) & 84.65~\small(0.6) & 87.13~\small(0.7) & - & - & - & - \\
    \textsc{BIC} \cite{lei2022bic} & 96.13~\small(0.9) & 96.94~\small(0.8) & 75.75~\small(1.3) & 79.24~\small(1.8) & - & - & - & - \\
    \textsc{BotBuster} \cite{ng2022botbuster} & 96.68~\small(0.2) & 96.42~\small(0.2) & 79.34~\small(0.8) & 82.47~\small(1.2) & - & - & - & - \\
    \textsc{GLNN}-RGT \cite{zhang2021graph} & 97.72~\small(0.7) & 98.22~\small(0.5) & 83.02~\small(0.5) & 85.60~\small(0.5) & - & - & - & - \\
    \textsc{GLNN}-BotRGCN \cite{zhang2021graph} & 97.71~\small(0.3) & 98.21~\small(0.2) & 82.56~\small(0.6) & 85.17~\small(0.4) & - & - & - & - \\
    \midrule[0.75pt]
    \textsc{RoBERTa}-finetune & 98.13~\small(0.3) & 97.43~\small(0.3)& 84.05~\small(0.2) & 86.13~\small(0.2) & 	96.32~\small(0.2) & 92.51~\small(0.5)  & 97.61~\small(0.4) & 98.58~\small(0.2) \\
    
    \midrule[0.75pt]
    \ourmethod{}-GNN/MLP (Ours)  & \textbf{99.06$^\ast$}~\small{(0.3)} & \textbf{99.26$^\ast$}~\small{(0.2)} & \underline{85.25}~\small{(0.2)} & \underline{87.38}~\small{(0.2)} & 
    \textbf{98.34$^\ast$}~\small{(0.3)} & \textbf{98.91$^\ast$}~\small{(0.2)} &
    \textbf{99.21$^\ast$}~\small{(0.1)} & \textbf{99.53$^\ast$}~\small{(0.1)}    \\
    \ourmethod{}-LM (Ours)  & \underline{98.74}~\small{(0.4)} & \underline{99.01}~\small{(0.3)} & \textbf{85.63$^\ast$}~\small{(0.2)} & \textbf{87.61$^\ast$}~\small{(0.3)} &
    \underline{98.28}~\small{(0.3)} & \underline{98.87}~\small{(0.2)} &
    \underline{99.21}~\small{(0.1)} & \underline{99.53}~\small{(0.1)}   \\
    \bottomrule[1.5pt]
\end{tabular}
\label{tab:results}
\end{table*}


\subsection{Main Results}
We evaluate our proposed \ourmethod{} and 10 other baselines on four Twitter bot detection benchmarks. Our results, presented in Table \ref{tab:results}, demonstrate that:
\begin{itemize}
    \item \ourmethod{} consistently outperforms all baseline methods on the four datasets with or without graph structure. Specifically, \ourmethod{} outperforms state-of-the-art 0.37\%, 0.97\%, 0.67\%, 0.40\% on Cresci-2015, TwiBot-20, Cresci-2015, Midterm-2018, respectively, even without graph structure in inference stage. Moreover, the performance of \textsc{RoBERTa}-finetune also confirms the effectiveness of \ourmethod{}.
    
    \item LM shows its core role in text-rich tasks. For datasets with graph structure (Cresci-2015,  and TwiBot-20), graph-based methods generally outperform non-graph-based methods. However, \ourmethod{} outperforms traditional graph-based methods. Moreover, no graph structure is needed during inference, demonstrating the efficiency of LMs. At the same time, we found that GNN enhanced by LM outperforms the original graph-based method (BotRGCN), indicating that LM can provide GNN with better initial representations. For datasets without graph structure, LM also shows better performance than traditional feature-based and text-based methods, demonstrating the important role of language models in text-rich tasks. 
    
    \item \ourmethod{} significantly outperforms GNN distillation baselines (GLNN \cite{zhang2021graph}). 
    The difference is that GLNN distills GNN into MLP, while \ourmethod{} designs an iterative strategy to distill GNN into LM. Our \ourmethod{}'s superiority indicates that in text-rich heterogeneous information networks, distilling GNN into LM is more effective for graph-less deployment. Moreover, the iterative distillation process enables GNN and LM to mutually enhance each other and therefore boost the performance of \ourmethod{}.
\end{itemize}

\subsection{Combination of Different GNNs and LMs}
\begin{figure}[t]
    \centering
    \includegraphics[width=0.8\linewidth]{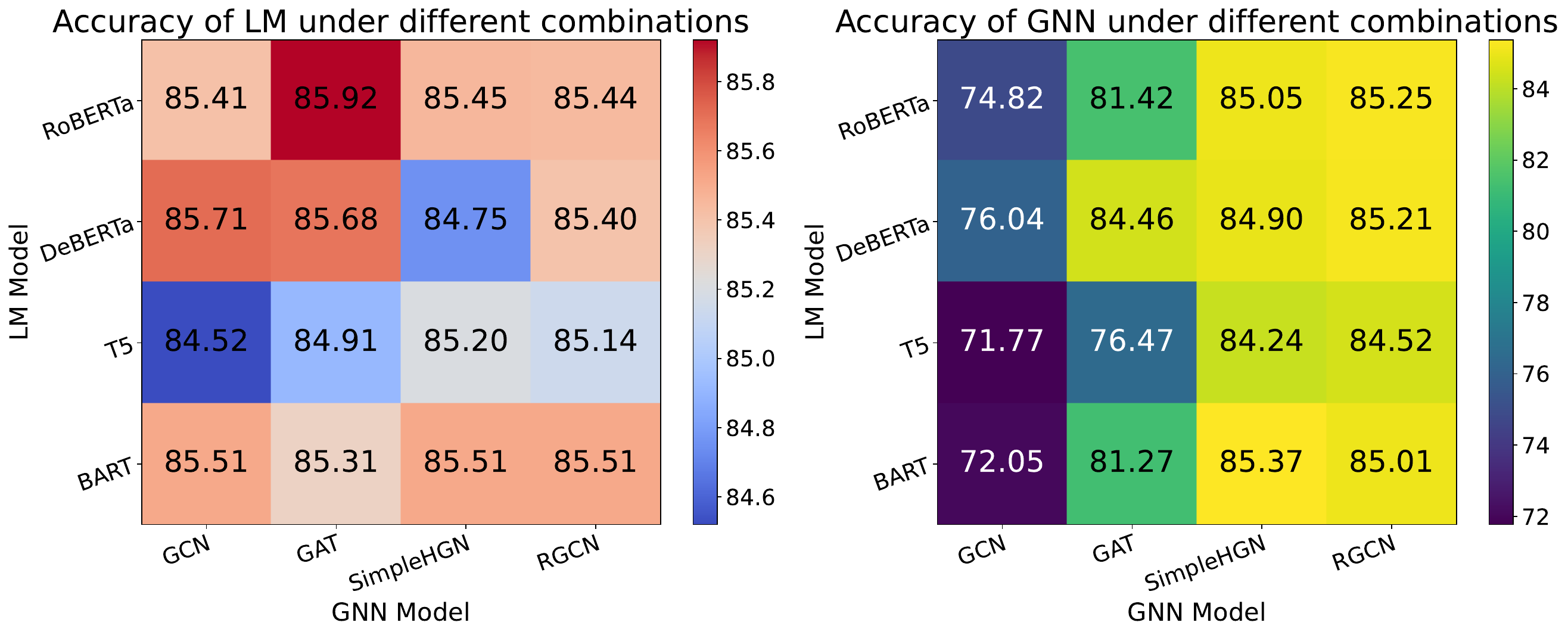}
    \caption{Accuracy with different combinations of LM and GNN. LMs are robust under different combinations while the performance of GNNs is highly dependent on the specific combination.}
    \label{fig:combination}
\end{figure}

Since various language models contain different initial inserted knowledge and inductive bias, while diverse GNNs could be applied to distill knowledge with different focuses. The effect of different combinations of GNN and LM is worth studying to show the \ourmethod{}'s robustness. Specifically, we selected four of each, resulting in $4 \times 4 = 16$ combinations. For LM, we selected T5 \cite{raffel2020exploring}, BART \cite{lewis2019bart}, RoBERTa \cite{liu2019roberta}, and DeBERTa \cite{he2020deberta}; for GNN, we selected GCN \cite{kipf2016semi}, GAT \cite{velivckovic2017graph}, SimpleHGN \cite{lv2021we}, and RGCN \cite{schlichtkrull2018modeling}. The experimental results on TwiBot-20 dataset are presented in Figure \ref{fig:combination}.

\begin{itemize}
    \item For LM, regardless of the GNN combined with it, it consistently achieves relatively stable and good performance, indicating that LM is not sensitive to GNN combined with it and can be compatible with various GNNs and present robust performance.
 Among them, the average performance (84.94) of T5 is slightly worse than the other three models, which are close to 85.5, indicating its slight deficiency of capacity to be adaptive to the current field.
    \item For GNNs, performance is sensitive to the choice of both LM and GNN. On the one hand, given LM, heterogeneous graph neural networks (SimpleHGN, RGCN) perform significantly better than homogeneous graph neural networks (GCN, GAT), indicating that graph structural information is very important for GNNs. On the other hand, given GNN, choosing a high-performance LM can improve the performance of GNN, which once again illustrates the core role of language models. The most obvious improvement can be seen in GAT, which achieves 84.46 when combined with DeBERTa, but only 76.47 when combined with T5. 
\end{itemize}

\subsection{Convergence Analysis} \label{sec:1}
\begin{figure}[t]
    \centering
    \subfigure[Convergence curve when LM is RoBERTa. When LM is finetuned, it takes fewer iterations to achieve optimal performance.]{\label{fig:convergence_RoBERTa}\includegraphics[width=0.8\linewidth]{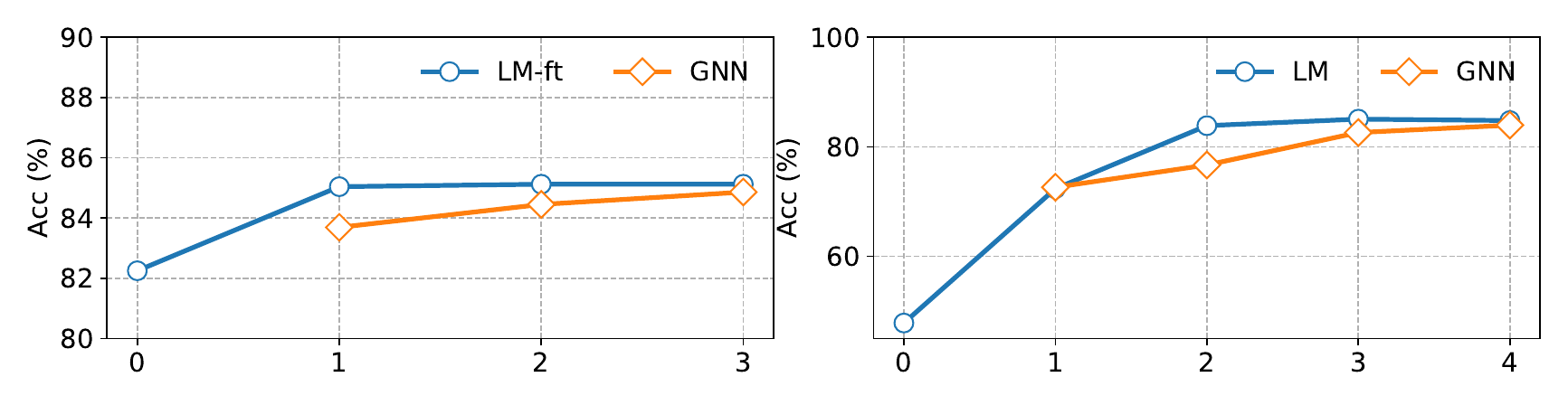}}
    \subfigure[Convergence curve when LM is T5. If LM is untuned, it cannot achieve optimal performance.]{\label{fig:convergence_T5}\includegraphics[width=0.8\linewidth]{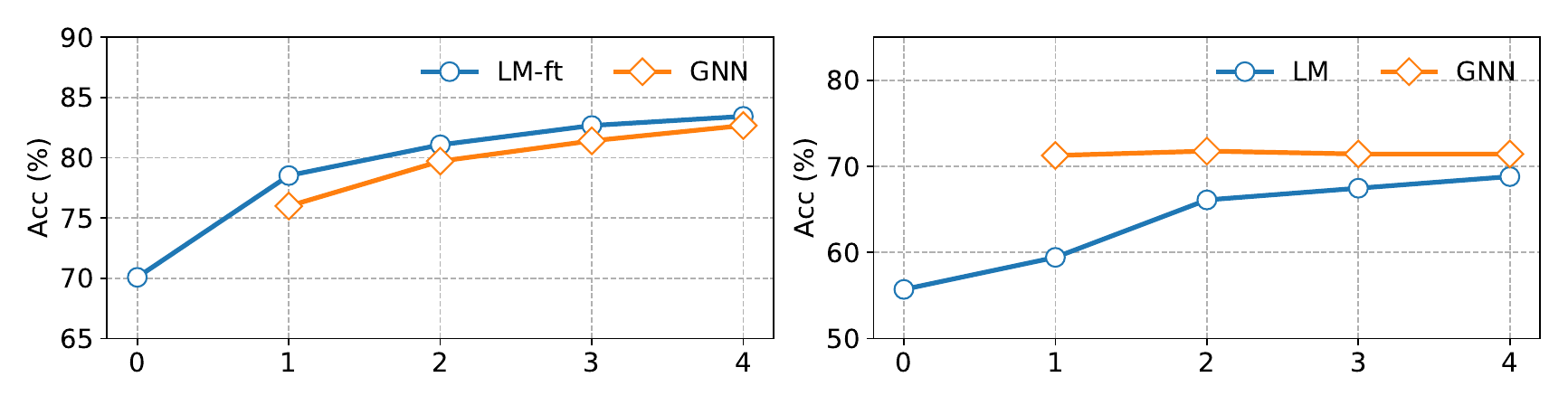}}
    \caption{Convergence curve when LM is finetuned and untuned. We confirm the necessity of domain adaptation for the language model's fast convergence speed and stability.}
    \label{fig:convergence}
\end{figure}

To examine the iterative training process and its impact on LM and GNN, respectively, we study how LM and GNN converge in \ourmethod{}. Specifically, we tried two settings on the TwiBot-20 dataset: \ourmethod{} under full settings and \ourmethod{} without LM domain adaptation to study their convergence process and the impact of LM domain adaptation. The convergence curves are shown in Figure \ref{fig:convergence_RoBERTa}. 

Two conclusions can be drawn from the curves: First, in the \ourmethod{} iteration, LM and GNN can promote each other's performance. In each iteration, both obtain improvements, indicating that LM generates better embeddings for GNN, and GNN also passes knowledge of graph structure to LM. LM can outperform GNN by fusing the inductive biases of GNN and its own inductive biases. Second, regardless of whether LM starts from a low or high starting point, it can converge to similar optimal performance, but the finetuned LM requires significantly fewer iterations to reach convergence than the untuned LM, which validates the effectiveness of LM's domain adaptation.

When replacing the LM with a lower-performing one (\emph{e.g.} T5) in Figure \ref{fig:convergence_T5}. It can be seen that only the finetuned LM can converge to the optimal value after a greater number of iterations than RoBERTa, while the untuned LM struggles to converge to the best performance.  This scenario further underscores the importance of fine-tuning LMs for adaptation.

\subsection{Prediction Consistency}
\begin{figure}[t]
    \centering
    \includegraphics[width=0.7\linewidth]{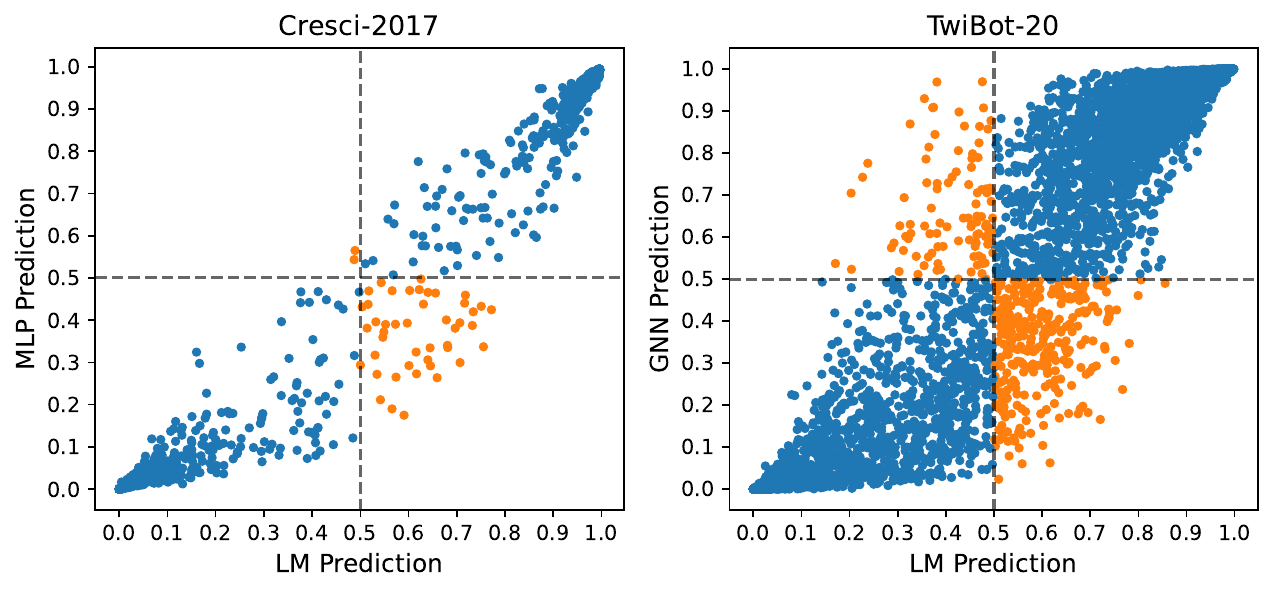}
    \caption{Predictions of LM and GNN/MLP on TwiBot-20 and Cresci-2017, where $x$ axis indicates the probability of LM predicting bot and $y$ axis indicates the probability of GNN/MLP predicting bot.}
    \label{fig:consistency}
\end{figure}

To explore the differences and similarities between the predictions of LM and GNN/MLP, we visualize the prediction of LM and GNN/MLP on TwiBot-20 and Cresci-2017 datasets in Figure \ref{fig:consistency}. Specifically, when the data points are in the first and third quadrants (colored in blue), it indicates that GNN/MLP and LM produce the same prediction; and the closer the data points are to $y=x$, the closer the prediction probabilities of GNN and LM/MLP are. The percentage of blue points among all points is 99.59\% on Cresci-2017 and 95.77\% on TwiBot-20, which suggests that LM and GNN/MLP produce consistent predictions on datasets with and without graph structure. The small differences in probability predictions can be regarded as the knowledge under the model's own inductive bias, which can be learned by another model through \ourmethod{} iteration.

\subsection{Data Efficiency Study}
Since manually annotating and collecting labels and graph structure information in the Twitter bot detection dataset requires massive human effort and time costs, a model that achieves high performance with fewer annotations and graph context is regarded as better. As a result, we study the influence of randomly removing some labels and local graph structures on model performance, in order to illustrate the capacity of our model in combating such data efficiency challenges. 

\begin{figure}[t]
    \centering
    \includegraphics[width=0.45\linewidth]{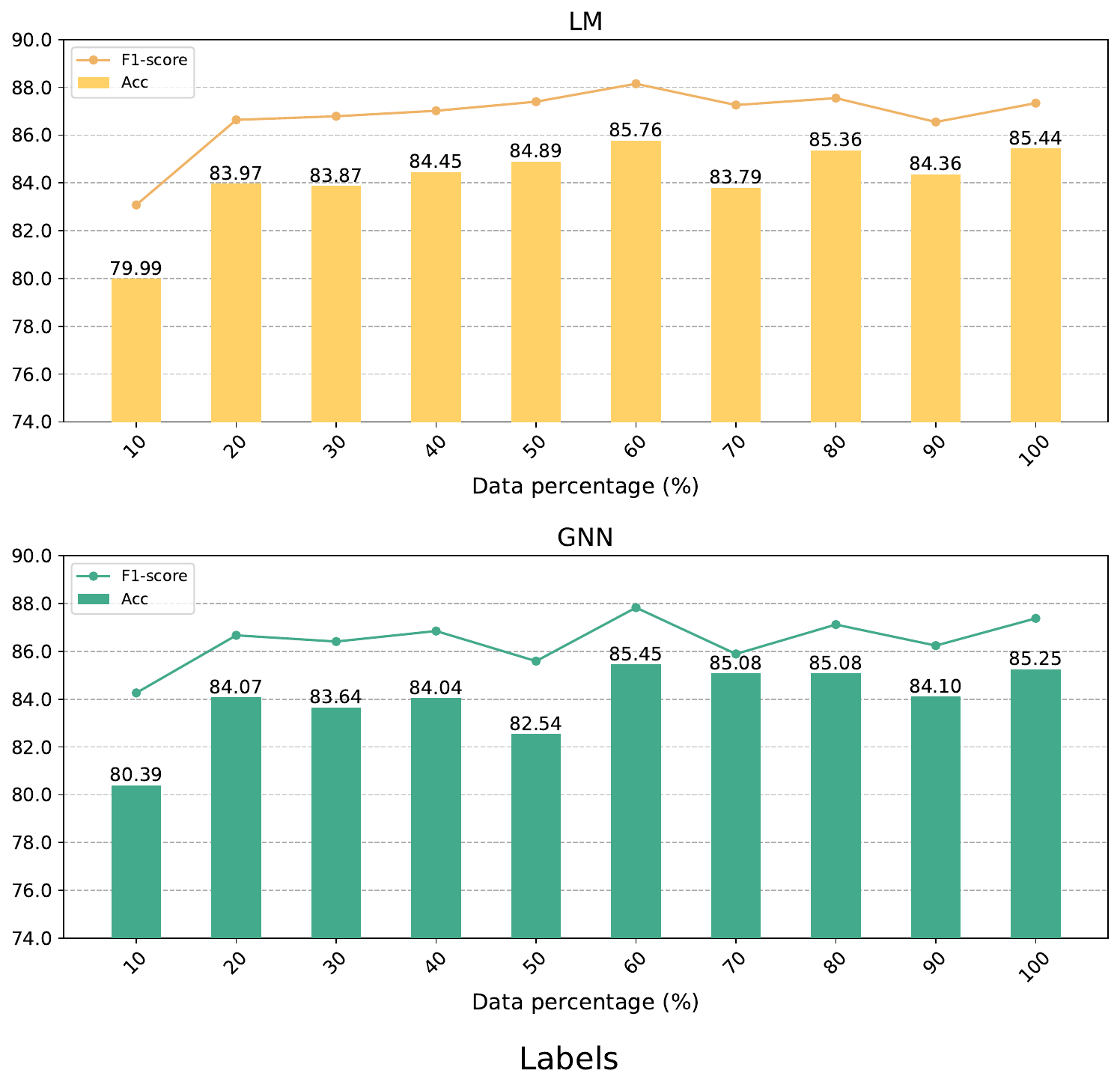}
    \includegraphics[width=0.45\linewidth]{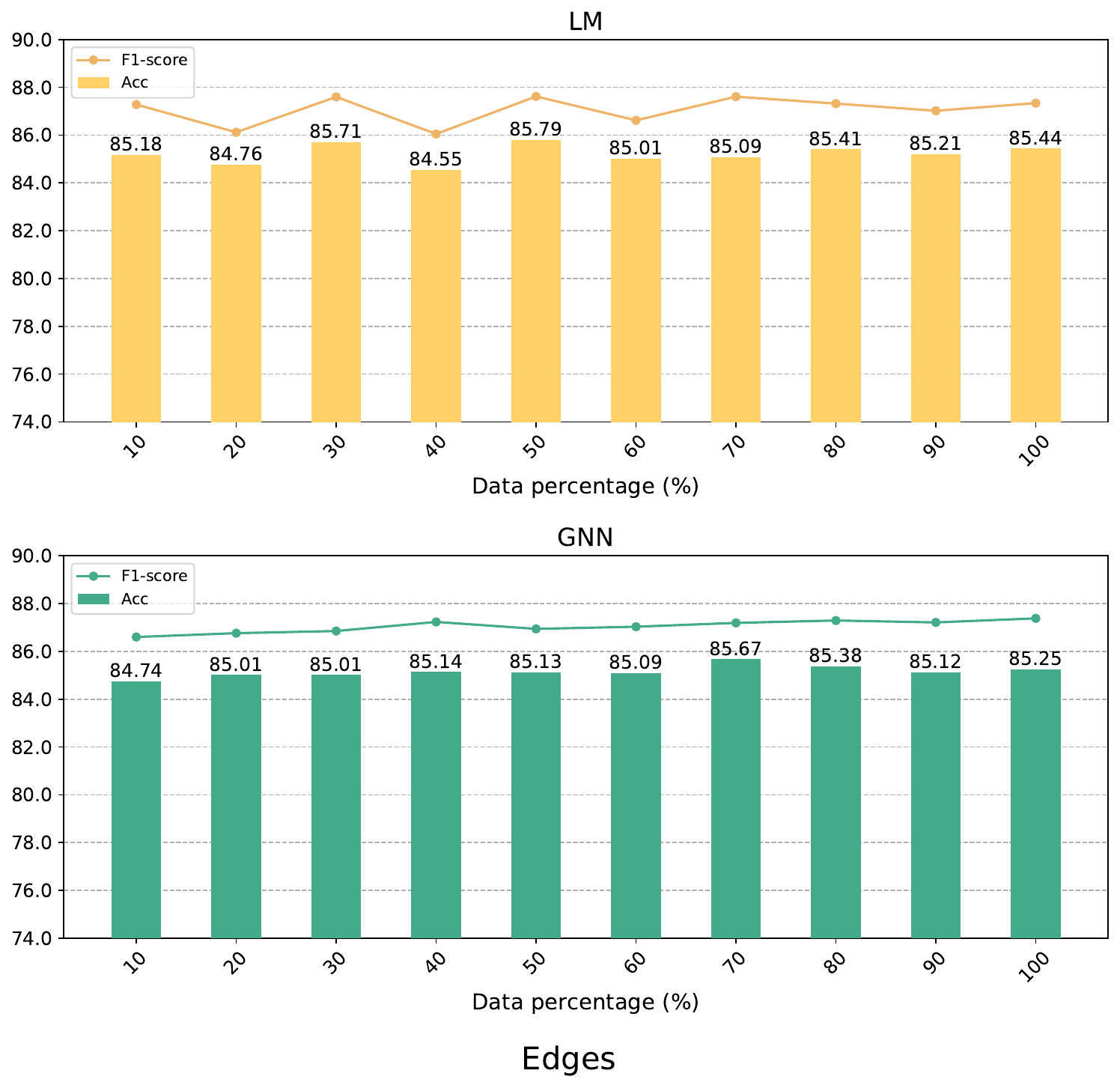}
    \caption{The performance of LM and GNN in \ourmethod{} with partial training set and graph structure. The left half is the label-level data efficiency experiment and the right half is the edge-level data efficiency experiment. It is demonstrated that \ourmethod{} remains highly robust even in the partial absence of graph structure and training labels.}
    \label{fig:data_efficiency}
\end{figure}

First, we trim the training set from 10\% to 100\% to conduct the label-level efficiency experiment. Then we randomly retained 10\% to 100\% of the edges in the original graph structure to conduct the edge-level efficiency experiment. The results are shown in Figure \ref{fig:data_efficiency}. It can be seen that \ourmethod{} has a relatively stable performance even with limited training labels. At the same time, it can be seen that \ourmethod{} is still very robust in the partial absence of graph structure, indicating the strong robustness of \ourmethod{} against limited annotations and contextual social network information.

\subsection{Ablation Study}

To study the impact of each module on the \ourmethod{}'s performance, we conduct ablation studies to validate our design choices. Specifically, we tried three different settings on the TwiBot-20 dataset: (1) remove the GNN module and \ourmethod{} iteration process, only LM finetuning; (2) remove LM finetuning and \ourmethod{} iteration process, only GNN training; (3) replace the GNN module with MLP and perform \ourmethod{} iteration, the results are shown in Table \ref{tab:ablation}.


\begin{table}[t]
    \renewcommand{\arraystretch}{0.8}
    \centering
    \caption{Ablation study of \ourmethod{} conducted on Twibot-20. Full model outperforms ablation models under different settings.}
    \begin{adjustbox}{max width=1\linewidth}
    \begin{tabular}{cccc}
    \toprule[1.5pt]
    \textbf{Category} & \textbf{Ablation Settings} & \textbf{Accuracy}& \textbf{Dif.}\\ 
    \midrule[0.75pt]
    full model & \ourmethod{} & \textbf{85.63}\small{(0.2)} & - \\
    \midrule[0.75pt]
    \multirow{2}[2]{*}{archtecture} &
        w/o GNN & 84.05\small{(0.2)}  &  -1.58\\
   
        & w/o LM &  73.97\small{(0.5)} & -11.28 \\

        \midrule[0.75pt]
   \multirow{2}[2]{*}{input} &
        w/o metadata & 71.34\small{(0.7)}  &  -14.29\\
        & w/o tweets & 82.80\small{(0.3)}  & -2.83 \\     
        \bottomrule[1.5pt]
    
    \end{tabular}
    \end{adjustbox}
    \label{tab:ablation}
\end{table}

To explore which part of the textual information is the most important to users, we remove the original three parts of textual information (metadata, description, tweets) respectively. We found that users' metadata has an extremely important influence, which is consistent with the findings in \cite{feng2021botrgcn}.

\section{Conclusion}
 Though graph-based methods achieve state-of-the-art Twitter bot detection performance, their inference depends on multi-hop neighbors which is time-consuming to fetch through API and may introduce sampling bias. Meanwhile, pretrained language models can achieve competitive performance without graph structure. Inspired by this finding and the necessity of social network structure information, we propose \ourmethod{}, a Twitter bot detection method that can iteratively distill social network contextual information in GNN into LM and subsequently do graph-free inference with LM to address data dependency problem. 
 Our experiments demonstrate that \ourmethod{} achieves state-of-the-art performance on four widely-used Twitter bot detection benchmarks. Extensive studies also show that \ourmethod{} is more robust, versatile, and efficient compared to previous state-of-the-art graph-based Twitter bot detection methods.

\section{Ethical Considerations}
\ourmethod{} is designed as an assistive tool rather than the sole decision maker in the Twitter bot detection application. Human judgment is essential for reaching final decisions when deployed on real social networks. First, \ourmethod{} can falsely identify legitimate accounts as bots, which may result in serious implications for individuals or organizations spreading crucial information. Second, both LM and GNN in \ourmethod{} may inherit the biases built into these models. For instance, LM can perpetuate stereotypes and social prejudices in pretraining data \cite{liang2021towards, nadeem2020stereoset} and GNN can discriminate against certain demographic groups \cite{dong2022edits}.

\section*{Acknowledgment}
This work was supported by the National Key Research and Development Program of China (No. 2022YFB3102600), National Nature Science Foundation of China (No. 62192781, No. 62272374, No. 62202367, No. 62250009, No. 62137002, No. 61937001), Innovative Research Group of the National Natural Science Foundation of China (61721002), Innovation Research Team of Ministry of Education (IRT\_17R86), Project of China Knowledge Center for Engineering Science and Technology, and Project of Chinese academy of engineering ``The Online and Offline Mixed Educational Service System for `The Belt and Road' Training in MOOC China''. We would like to express our gratitude for the support of K. C. Wong Education Foundation. We also appreciate the reviewers and chairs for their constructive feedback. Lastly, we would like to thank all LUD lab members for fostering a collaborative research environment.

\bibliographystyle{ACM-Reference-Format}
\balance
\bibliography{software}


\begin{thebibliography}{56}


\ifx \showCODEN    \undefined \def \showCODEN     #1{\unskip}     \fi
\ifx \showDOI      \undefined \def \showDOI       #1{#1}\fi
\ifx \showISBNx    \undefined \def \showISBNx     #1{\unskip}     \fi
\ifx \showISBNxiii \undefined \def \showISBNxiii  #1{\unskip}     \fi
\ifx \showISSN     \undefined \def \showISSN      #1{\unskip}     \fi
\ifx \showLCCN     \undefined \def \showLCCN      #1{\unskip}     \fi
\ifx \shownote     \undefined \def \shownote      #1{#1}          \fi
\ifx \showarticletitle \undefined \def \showarticletitle #1{#1}   \fi
\ifx \showURL      \undefined \def \showURL       {\relax}        \fi
\providecommand\bibfield[2]{#2}
\providecommand\bibinfo[2]{#2}
\providecommand\natexlab[1]{#1}
\providecommand\showeprint[2][]{arXiv:#2}

\bibitem[Abreu et~al\mbox{.}(2020)]%
        {abreu2020twitter}
\bibfield{author}{\bibinfo{person}{Jefferson Viana~Fonseca Abreu},
  \bibinfo{person}{C{\'e}lia~Ghedini Ralha}, {and} \bibinfo{person}{Jo{\~a}o
  Jos{\'e}~Costa Gondim}.} \bibinfo{year}{2020}\natexlab{}.
\newblock \showarticletitle{Twitter bot detection with reduced feature set}. In
  \bibinfo{booktitle}{\emph{2020 IEEE International Conference on Intelligence
  and Security Informatics (ISI)}}. IEEE, \bibinfo{pages}{1--6}.
\newblock


\bibitem[Ali~Alhosseini et~al\mbox{.}(2019)]%
        {ali2019detect}
\bibfield{author}{\bibinfo{person}{Seyed Ali~Alhosseini}, \bibinfo{person}{Raad
  Bin~Tareaf}, \bibinfo{person}{Pejman Najafi}, {and}
  \bibinfo{person}{Christoph Meinel}.} \bibinfo{year}{2019}\natexlab{}.
\newblock \showarticletitle{Detect me if you can: Spam bot detection using
  inductive representation learning}. In \bibinfo{booktitle}{\emph{Companion
  Proceedings of The 2019 World Wide Web Conference}}.
  \bibinfo{pages}{148--153}.
\newblock


\bibitem[Beskow and Carley(2018)]%
        {beskow2018bot}
\bibfield{author}{\bibinfo{person}{David~M Beskow} {and}
  \bibinfo{person}{Kathleen~M Carley}.} \bibinfo{year}{2018}\natexlab{}.
\newblock \showarticletitle{Bot-hunter: a tiered approach to detecting \&
  characterizing automated activity on twitter}. In
  \bibinfo{booktitle}{\emph{Conference paper. SBP-BRiMS: International
  conference on social computing, behavioral-cultural modeling and prediction
  and behavior representation in modeling and simulation}},
  Vol.~\bibinfo{volume}{3}.
\newblock


\bibitem[Cresci et~al\mbox{.}(2015)]%
        {cresci2015fame}
\bibfield{author}{\bibinfo{person}{Stefano Cresci}, \bibinfo{person}{Roberto
  Di~Pietro}, \bibinfo{person}{Marinella Petrocchi}, \bibinfo{person}{Angelo
  Spognardi}, {and} \bibinfo{person}{Maurizio Tesconi}.}
  \bibinfo{year}{2015}\natexlab{}.
\newblock \showarticletitle{Fame for sale: Efficient detection of fake Twitter
  followers}.
\newblock \bibinfo{journal}{\emph{Decision Support Systems}}
  \bibinfo{volume}{80} (\bibinfo{year}{2015}), \bibinfo{pages}{56--71}.
\newblock


\bibitem[Cresci et~al\mbox{.}(2017)]%
        {cresci2017paradigm}
\bibfield{author}{\bibinfo{person}{Stefano Cresci}, \bibinfo{person}{Roberto
  Di~Pietro}, \bibinfo{person}{Marinella Petrocchi}, \bibinfo{person}{Angelo
  Spognardi}, {and} \bibinfo{person}{Maurizio Tesconi}.}
  \bibinfo{year}{2017}\natexlab{}.
\newblock \showarticletitle{The paradigm-shift of social spambots: Evidence,
  theories, and tools for the arms race}. In
  \bibinfo{booktitle}{\emph{Proceedings of the 26th international conference on
  world wide web companion}}. \bibinfo{pages}{963--972}.
\newblock


\bibitem[Dong et~al\mbox{.}(2022)]%
        {dong2022edits}
\bibfield{author}{\bibinfo{person}{Yushun Dong}, \bibinfo{person}{Ninghao Liu},
  \bibinfo{person}{Brian Jalaian}, {and} \bibinfo{person}{Jundong Li}.}
  \bibinfo{year}{2022}\natexlab{}.
\newblock \showarticletitle{Edits: Modeling and mitigating data bias for graph
  neural networks}. In \bibinfo{booktitle}{\emph{Proceedings of the ACM Web
  Conference 2022}}. \bibinfo{pages}{1259--1269}.
\newblock


\bibitem[Echeverr{\"\i}{\pounds}!`~a et~al\mbox{.}(2018)]%
        {echeverri2018lobo}
\bibfield{author}{\bibinfo{person}{Juan Echeverr{\"\i}{\pounds}!`~a},
  \bibinfo{person}{Emiliano De~Cristofaro}, \bibinfo{person}{Nicolas
  Kourtellis}, \bibinfo{person}{Ilias Leontiadis}, \bibinfo{person}{Gianluca
  Stringhini}, {and} \bibinfo{person}{Shi Zhou}.}
  \bibinfo{year}{2018}\natexlab{}.
\newblock \showarticletitle{LOBO: Evaluation of generalization deficiencies in
  Twitter bot classifiers}. In \bibinfo{booktitle}{\emph{Proceedings of the
  34th annual computer security applications conference}}.
  \bibinfo{pages}{137--146}.
\newblock


\bibitem[Feng et~al\mbox{.}(2022a)]%
        {feng2022heterogeneity}
\bibfield{author}{\bibinfo{person}{Shangbin Feng}, \bibinfo{person}{Zhaoxuan
  Tan}, \bibinfo{person}{Rui Li}, {and} \bibinfo{person}{Minnan Luo}.}
  \bibinfo{year}{2022}\natexlab{a}.
\newblock \showarticletitle{Heterogeneity-aware twitter bot detection with
  relational graph transformers}. In \bibinfo{booktitle}{\emph{Proceedings of
  the AAAI Conference on Artificial Intelligence}}, Vol.~\bibinfo{volume}{36}.
  \bibinfo{pages}{3977--3985}.
\newblock


\bibitem[Feng et~al\mbox{.}(2022b)]%
        {feng2022twibot}
\bibfield{author}{\bibinfo{person}{Shangbin Feng}, \bibinfo{person}{Zhaoxuan
  Tan}, \bibinfo{person}{Herun Wan}, \bibinfo{person}{Ningnan Wang},
  \bibinfo{person}{Zilong Chen}, \bibinfo{person}{Binchi Zhang},
  \bibinfo{person}{Qinghua Zheng}, \bibinfo{person}{Wenqian Zhang},
  \bibinfo{person}{Zhenyu Lei}, \bibinfo{person}{Shujie Yang}, {et~al\mbox{.}}}
  \bibinfo{year}{2022}\natexlab{b}.
\newblock \showarticletitle{TwiBot-22: Towards graph-based Twitter bot
  detection}.
\newblock \bibinfo{journal}{\emph{arXiv preprint arXiv:2206.04564}}
  (\bibinfo{year}{2022}).
\newblock


\bibitem[Feng et~al\mbox{.}(2021b)]%
        {feng2021satar}
\bibfield{author}{\bibinfo{person}{Shangbin Feng}, \bibinfo{person}{Herun Wan},
  \bibinfo{person}{Ningnan Wang}, \bibinfo{person}{Jundong Li}, {and}
  \bibinfo{person}{Minnan Luo}.} \bibinfo{year}{2021}\natexlab{b}.
\newblock \showarticletitle{Satar: A self-supervised approach to twitter
  account representation learning and its application in bot detection}. In
  \bibinfo{booktitle}{\emph{Proceedings of the 30th ACM International
  Conference on Information \& Knowledge Management}}.
  \bibinfo{pages}{3808--3817}.
\newblock


\bibitem[Feng et~al\mbox{.}(2021c)]%
        {feng2021twibot}
\bibfield{author}{\bibinfo{person}{Shangbin Feng}, \bibinfo{person}{Herun Wan},
  \bibinfo{person}{Ningnan Wang}, \bibinfo{person}{Jundong Li}, {and}
  \bibinfo{person}{Minnan Luo}.} \bibinfo{year}{2021}\natexlab{c}.
\newblock \showarticletitle{Twibot-20: A comprehensive twitter bot detection
  benchmark}. In \bibinfo{booktitle}{\emph{Proceedings of the 30th ACM
  International Conference on Information \& Knowledge Management}}.
  \bibinfo{pages}{4485--4494}.
\newblock


\bibitem[Feng et~al\mbox{.}(2021a)]%
        {feng2021botrgcn}
\bibfield{author}{\bibinfo{person}{Shangbin Feng}, \bibinfo{person}{Herun Wan},
  \bibinfo{person}{Ningnan Wang}, {and} \bibinfo{person}{Minnan Luo}.}
  \bibinfo{year}{2021}\natexlab{a}.
\newblock \showarticletitle{BotRGCN: Twitter bot detection with relational
  graph convolutional networks}. In \bibinfo{booktitle}{\emph{Proceedings of
  the 2021 IEEE/ACM International Conference on Advances in Social Networks
  Analysis and Mining}}. \bibinfo{pages}{236--239}.
\newblock


\bibitem[Ferrara et~al\mbox{.}(2016)]%
        {ferrara2016rise}
\bibfield{author}{\bibinfo{person}{Emilio Ferrara}, \bibinfo{person}{Onur
  Varol}, \bibinfo{person}{Clayton Davis}, \bibinfo{person}{Filippo Menczer},
  {and} \bibinfo{person}{Alessandro Flammini}.}
  \bibinfo{year}{2016}\natexlab{}.
\newblock \showarticletitle{The rise of social bots}.
\newblock \bibinfo{journal}{\emph{Commun. ACM}} \bibinfo{volume}{59},
  \bibinfo{number}{7} (\bibinfo{year}{2016}), \bibinfo{pages}{96--104}.
\newblock


\bibitem[Fey and Lenssen(2019)]%
        {fey2019fast}
\bibfield{author}{\bibinfo{person}{Matthias Fey} {and}
  \bibinfo{person}{Jan~Eric Lenssen}.} \bibinfo{year}{2019}\natexlab{}.
\newblock \showarticletitle{Fast graph representation learning with PyTorch
  Geometric}.
\newblock \bibinfo{journal}{\emph{arXiv preprint arXiv:1903.02428}}
  (\bibinfo{year}{2019}).
\newblock


\bibitem[Guo et~al\mbox{.}(2021)]%
        {guo2021social}
\bibfield{author}{\bibinfo{person}{Qinglang Guo}, \bibinfo{person}{Haiyong
  Xie}, \bibinfo{person}{Yangyang Li}, \bibinfo{person}{Wen Ma}, {and}
  \bibinfo{person}{Chao Zhang}.} \bibinfo{year}{2021}\natexlab{}.
\newblock \showarticletitle{Social bots detection via fusing bert and graph
  convolutional networks}.
\newblock \bibinfo{journal}{\emph{Symmetry}} \bibinfo{volume}{14},
  \bibinfo{number}{1} (\bibinfo{year}{2021}), \bibinfo{pages}{30}.
\newblock


\bibitem[Guo et~al\mbox{.}(2022)]%
        {guo2022linkless}
\bibfield{author}{\bibinfo{person}{Zhichun Guo}, \bibinfo{person}{William
  Shiao}, \bibinfo{person}{Shichang Zhang}, \bibinfo{person}{Yozen Liu},
  \bibinfo{person}{Nitesh Chawla}, \bibinfo{person}{Neil Shah}, {and}
  \bibinfo{person}{Tong Zhao}.} \bibinfo{year}{2022}\natexlab{}.
\newblock \showarticletitle{Linkless Link Prediction via Relational
  Distillation}.
\newblock \bibinfo{journal}{\emph{arXiv preprint arXiv:2210.05801}}
  (\bibinfo{year}{2022}).
\newblock


\bibitem[Hayawi et~al\mbox{.}(2022)]%
        {hayawi2022deeprobot}
\bibfield{author}{\bibinfo{person}{Kadhim Hayawi}, \bibinfo{person}{Sujith
  Mathew}, \bibinfo{person}{Neethu Venugopal}, \bibinfo{person}{Mohammad~M
  Masud}, {and} \bibinfo{person}{Pin-Han Ho}.} \bibinfo{year}{2022}\natexlab{}.
\newblock \showarticletitle{DeeProBot: a hybrid deep neural network model for
  social bot detection based on user profile data}.
\newblock \bibinfo{journal}{\emph{Social Network Analysis and Mining}}
  \bibinfo{volume}{12}, \bibinfo{number}{1} (\bibinfo{year}{2022}),
  \bibinfo{pages}{43}.
\newblock


\bibitem[He et~al\mbox{.}(2020)]%
        {he2020deberta}
\bibfield{author}{\bibinfo{person}{Pengcheng He}, \bibinfo{person}{Xiaodong
  Liu}, \bibinfo{person}{Jianfeng Gao}, {and} \bibinfo{person}{Weizhu Chen}.}
  \bibinfo{year}{2020}\natexlab{}.
\newblock \showarticletitle{Deberta: Decoding-enhanced bert with disentangled
  attention}.
\newblock \bibinfo{journal}{\emph{arXiv preprint arXiv:2006.03654}}
  (\bibinfo{year}{2020}).
\newblock


\bibitem[Hinton et~al\mbox{.}(2015)]%
        {hinton2015distilling}
\bibfield{author}{\bibinfo{person}{Geoffrey Hinton}, \bibinfo{person}{Oriol
  Vinyals}, {and} \bibinfo{person}{Jeff Dean}.}
  \bibinfo{year}{2015}\natexlab{}.
\newblock \showarticletitle{Distilling the knowledge in a neural network}.
\newblock \bibinfo{journal}{\emph{arXiv preprint arXiv:1503.02531}}
  (\bibinfo{year}{2015}).
\newblock


\bibitem[Howard et~al\mbox{.}(2017)]%
        {howard2017junk}
\bibfield{author}{\bibinfo{person}{Philip~N Howard}, \bibinfo{person}{Gillian
  Bolsover}, \bibinfo{person}{Bence Kollanyi}, \bibinfo{person}{Samantha
  Bradshaw}, {and} \bibinfo{person}{Lisa-Maria Neudert}.}
  \bibinfo{year}{2017}\natexlab{}.
\newblock \showarticletitle{Junk news and bots during the US election: What
  were Michigan voters sharing over Twitter}.
\newblock \bibinfo{journal}{\emph{CompProp, OII, Data Memo}}
  \bibinfo{volume}{1} (\bibinfo{year}{2017}).
\newblock


\bibitem[Jiao et~al\mbox{.}(2019)]%
        {jiao2019tinybert}
\bibfield{author}{\bibinfo{person}{Xiaoqi Jiao}, \bibinfo{person}{Yichun Yin},
  \bibinfo{person}{Lifeng Shang}, \bibinfo{person}{Xin Jiang},
  \bibinfo{person}{Xiao Chen}, \bibinfo{person}{Linlin Li},
  \bibinfo{person}{Fang Wang}, {and} \bibinfo{person}{Qun Liu}.}
  \bibinfo{year}{2019}\natexlab{}.
\newblock \showarticletitle{Tinybert: Distilling bert for natural language
  understanding}.
\newblock \bibinfo{journal}{\emph{arXiv preprint arXiv:1909.10351}}
  (\bibinfo{year}{2019}).
\newblock


\bibitem[Kipf and Welling(2016)]%
        {kipf2016semi}
\bibfield{author}{\bibinfo{person}{Thomas~N Kipf} {and} \bibinfo{person}{Max
  Welling}.} \bibinfo{year}{2016}\natexlab{}.
\newblock \showarticletitle{Semi-supervised classification with graph
  convolutional networks}.
\newblock \bibinfo{journal}{\emph{arXiv preprint arXiv:1609.02907}}
  (\bibinfo{year}{2016}).
\newblock


\bibitem[Kouvela et~al\mbox{.}(2020)]%
        {kouvela2020bot}
\bibfield{author}{\bibinfo{person}{Maria Kouvela}, \bibinfo{person}{Ilias
  Dimitriadis}, {and} \bibinfo{person}{Athena Vakali}.}
  \bibinfo{year}{2020}\natexlab{}.
\newblock \showarticletitle{Bot-Detective: An explainable Twitter bot detection
  service with crowdsourcing functionalities}. In
  \bibinfo{booktitle}{\emph{Proceedings of the 12th International Conference on
  Management of Digital EcoSystems}}. \bibinfo{pages}{55--63}.
\newblock


\bibitem[Kudugunta and Ferrara(2018)]%
        {kudugunta2018deep}
\bibfield{author}{\bibinfo{person}{Sneha Kudugunta} {and}
  \bibinfo{person}{Emilio Ferrara}.} \bibinfo{year}{2018}\natexlab{}.
\newblock \showarticletitle{Deep neural networks for bot detection}.
\newblock \bibinfo{journal}{\emph{Information Sciences}}  \bibinfo{volume}{467}
  (\bibinfo{year}{2018}), \bibinfo{pages}{312--322}.
\newblock


\bibitem[Kullback and Leibler(1951)]%
        {kullback1951information}
\bibfield{author}{\bibinfo{person}{Solomon Kullback} {and}
  \bibinfo{person}{Richard~A Leibler}.} \bibinfo{year}{1951}\natexlab{}.
\newblock \showarticletitle{On information and sufficiency}.
\newblock \bibinfo{journal}{\emph{The annals of mathematical statistics}}
  \bibinfo{volume}{22}, \bibinfo{number}{1} (\bibinfo{year}{1951}),
  \bibinfo{pages}{79--86}.
\newblock


\bibitem[Lei et~al\mbox{.}(2022)]%
        {lei2022bic}
\bibfield{author}{\bibinfo{person}{Zhenyu Lei}, \bibinfo{person}{Herun Wan},
  \bibinfo{person}{Wenqian Zhang}, \bibinfo{person}{Shangbin Feng},
  \bibinfo{person}{Zilong Chen}, \bibinfo{person}{Qinghua Zheng}, {and}
  \bibinfo{person}{Minnan Luo}.} \bibinfo{year}{2022}\natexlab{}.
\newblock \showarticletitle{BIC: Twitter Bot Detection with Text-Graph
  Interaction and Semantic Consistency}.
\newblock \bibinfo{journal}{\emph{arXiv preprint arXiv:2208.08320}}
  (\bibinfo{year}{2022}).
\newblock


\bibitem[Leskovec and Faloutsos(2006)]%
        {leskovec2006sampling}
\bibfield{author}{\bibinfo{person}{Jure Leskovec} {and}
  \bibinfo{person}{Christos Faloutsos}.} \bibinfo{year}{2006}\natexlab{}.
\newblock \showarticletitle{Sampling from large graphs}. In
  \bibinfo{booktitle}{\emph{Proceedings of the 12th ACM SIGKDD international
  conference on Knowledge discovery and data mining}}.
  \bibinfo{pages}{631--636}.
\newblock


\bibitem[Lewis et~al\mbox{.}(2019)]%
        {lewis2019bart}
\bibfield{author}{\bibinfo{person}{Mike Lewis}, \bibinfo{person}{Yinhan Liu},
  \bibinfo{person}{Naman Goyal}, \bibinfo{person}{Marjan Ghazvininejad},
  \bibinfo{person}{Abdelrahman Mohamed}, \bibinfo{person}{Omer Levy},
  \bibinfo{person}{Ves Stoyanov}, {and} \bibinfo{person}{Luke Zettlemoyer}.}
  \bibinfo{year}{2019}\natexlab{}.
\newblock \showarticletitle{Bart: Denoising sequence-to-sequence pre-training
  for natural language generation, translation, and comprehension}.
\newblock \bibinfo{journal}{\emph{arXiv preprint arXiv:1910.13461}}
  (\bibinfo{year}{2019}).
\newblock


\bibitem[Liang et~al\mbox{.}(2021)]%
        {liang2021towards}
\bibfield{author}{\bibinfo{person}{Paul~Pu Liang}, \bibinfo{person}{Chiyu Wu},
  \bibinfo{person}{Louis-Philippe Morency}, {and} \bibinfo{person}{Ruslan
  Salakhutdinov}.} \bibinfo{year}{2021}\natexlab{}.
\newblock \showarticletitle{Towards understanding and mitigating social biases
  in language models}. In \bibinfo{booktitle}{\emph{International Conference on
  Machine Learning}}. PMLR, \bibinfo{pages}{6565--6576}.
\newblock


\bibitem[Liu et~al\mbox{.}(2019)]%
        {liu2019roberta}
\bibfield{author}{\bibinfo{person}{Yinhan Liu}, \bibinfo{person}{Myle Ott},
  \bibinfo{person}{Naman Goyal}, \bibinfo{person}{Jingfei Du},
  \bibinfo{person}{Mandar Joshi}, \bibinfo{person}{Danqi Chen},
  \bibinfo{person}{Omer Levy}, \bibinfo{person}{Mike Lewis},
  \bibinfo{person}{Luke Zettlemoyer}, {and} \bibinfo{person}{Veselin
  Stoyanov}.} \bibinfo{year}{2019}\natexlab{}.
\newblock \showarticletitle{Roberta: A robustly optimized bert pretraining
  approach}.
\newblock \bibinfo{journal}{\emph{arXiv preprint arXiv:1907.11692}}
  (\bibinfo{year}{2019}).
\newblock


\bibitem[Liu et~al\mbox{.}(2023)]%
        {liu2023botmoe}
\bibfield{author}{\bibinfo{person}{Yuhan Liu}, \bibinfo{person}{Zhaoxuan Tan},
  \bibinfo{person}{Heng Wang}, \bibinfo{person}{Shangbin Feng},
  \bibinfo{person}{Qinghua Zheng}, {and} \bibinfo{person}{Minnan Luo}.}
  \bibinfo{year}{2023}\natexlab{}.
\newblock \showarticletitle{BotMoE: Twitter Bot Detection with Community-Aware
  Mixtures of Modal-Specific Experts}.
\newblock \bibinfo{journal}{\emph{arXiv preprint arXiv:2304.06280}}
  (\bibinfo{year}{2023}).
\newblock


\bibitem[Lv et~al\mbox{.}(2021)]%
        {lv2021we}
\bibfield{author}{\bibinfo{person}{Qingsong Lv}, \bibinfo{person}{Ming Ding},
  \bibinfo{person}{Qiang Liu}, \bibinfo{person}{Yuxiang Chen},
  \bibinfo{person}{Wenzheng Feng}, \bibinfo{person}{Siming He},
  \bibinfo{person}{Chang Zhou}, \bibinfo{person}{Jianguo Jiang},
  \bibinfo{person}{Yuxiao Dong}, {and} \bibinfo{person}{Jie Tang}.}
  \bibinfo{year}{2021}\natexlab{}.
\newblock \showarticletitle{Are we really making much progress? revisiting,
  benchmarking and refining heterogeneous graph neural networks}. In
  \bibinfo{booktitle}{\emph{Proceedings of the 27th ACM SIGKDD conference on
  knowledge discovery \& data mining}}. \bibinfo{pages}{1150--1160}.
\newblock


\bibitem[Miller et~al\mbox{.}(2014)]%
        {miller2014twitter}
\bibfield{author}{\bibinfo{person}{Zachary Miller}, \bibinfo{person}{Brian
  Dickinson}, \bibinfo{person}{William Deitrick}, \bibinfo{person}{Wei Hu},
  {and} \bibinfo{person}{Alex~Hai Wang}.} \bibinfo{year}{2014}\natexlab{}.
\newblock \showarticletitle{Twitter spammer detection using data stream
  clustering}.
\newblock \bibinfo{journal}{\emph{Information Sciences}}  \bibinfo{volume}{260}
  (\bibinfo{year}{2014}), \bibinfo{pages}{64--73}.
\newblock


\bibitem[Nadeem et~al\mbox{.}(2020)]%
        {nadeem2020stereoset}
\bibfield{author}{\bibinfo{person}{Moin Nadeem}, \bibinfo{person}{Anna Bethke},
  {and} \bibinfo{person}{Siva Reddy}.} \bibinfo{year}{2020}\natexlab{}.
\newblock \showarticletitle{StereoSet: Measuring stereotypical bias in
  pretrained language models}.
\newblock \bibinfo{journal}{\emph{arXiv preprint arXiv:2004.09456}}
  (\bibinfo{year}{2020}).
\newblock


\bibitem[Ng and Carley(2022)]%
        {ng2022botbuster}
\bibfield{author}{\bibinfo{person}{Lynnette Hui~Xian Ng} {and}
  \bibinfo{person}{Kathleen~M Carley}.} \bibinfo{year}{2022}\natexlab{}.
\newblock \showarticletitle{BotBuster: Multi-platform Bot Detection Using A
  Mixture of Experts}.
\newblock \bibinfo{journal}{\emph{arXiv preprint arXiv:2207.13658}}
  (\bibinfo{year}{2022}).
\newblock


\bibitem[Nguyen et~al\mbox{.}(2020)]%
        {nguyen2020bertweet}
\bibfield{author}{\bibinfo{person}{Dat~Quoc Nguyen}, \bibinfo{person}{Thanh
  Vu}, {and} \bibinfo{person}{Anh~Tuan Nguyen}.}
  \bibinfo{year}{2020}\natexlab{}.
\newblock \showarticletitle{BERTweet: A pre-trained language model for English
  Tweets}.
\newblock \bibinfo{journal}{\emph{arXiv preprint arXiv:2005.10200}}
  (\bibinfo{year}{2020}).
\newblock


\bibitem[Paszke et~al\mbox{.}(2019)]%
        {paszke2019pytorch}
\bibfield{author}{\bibinfo{person}{Adam Paszke}, \bibinfo{person}{Sam Gross},
  \bibinfo{person}{Francisco Massa}, \bibinfo{person}{Adam Lerer},
  \bibinfo{person}{James Bradbury}, \bibinfo{person}{Gregory Chanan},
  \bibinfo{person}{Trevor Killeen}, \bibinfo{person}{Zeming Lin},
  \bibinfo{person}{Natalia Gimelshein}, \bibinfo{person}{Luca Antiga},
  {et~al\mbox{.}}} \bibinfo{year}{2019}\natexlab{}.
\newblock \showarticletitle{Pytorch: An imperative style, high-performance deep
  learning library}.
\newblock \bibinfo{journal}{\emph{Advances in neural information processing
  systems}}  \bibinfo{volume}{32} (\bibinfo{year}{2019}).
\newblock


\bibitem[Pedregosa et~al\mbox{.}(2011)]%
        {pedregosa2011scikit}
\bibfield{author}{\bibinfo{person}{Fabian Pedregosa}, \bibinfo{person}{Ga{\"e}l
  Varoquaux}, \bibinfo{person}{Alexandre Gramfort}, \bibinfo{person}{Vincent
  Michel}, \bibinfo{person}{Bertrand Thirion}, \bibinfo{person}{Olivier
  Grisel}, \bibinfo{person}{Mathieu Blondel}, \bibinfo{person}{Peter
  Prettenhofer}, \bibinfo{person}{Ron Weiss}, \bibinfo{person}{Vincent
  Dubourg}, {et~al\mbox{.}}} \bibinfo{year}{2011}\natexlab{}.
\newblock \showarticletitle{Scikit-learn: Machine learning in Python}.
\newblock \bibinfo{journal}{\emph{the Journal of machine Learning research}}
  \bibinfo{volume}{12} (\bibinfo{year}{2011}), \bibinfo{pages}{2825--2830}.
\newblock


\bibitem[Raffel et~al\mbox{.}(2020)]%
        {raffel2020exploring}
\bibfield{author}{\bibinfo{person}{Colin Raffel}, \bibinfo{person}{Noam
  Shazeer}, \bibinfo{person}{Adam Roberts}, \bibinfo{person}{Katherine Lee},
  \bibinfo{person}{Sharan Narang}, \bibinfo{person}{Michael Matena},
  \bibinfo{person}{Yanqi Zhou}, \bibinfo{person}{Wei Li}, {and}
  \bibinfo{person}{Peter~J Liu}.} \bibinfo{year}{2020}\natexlab{}.
\newblock \showarticletitle{Exploring the limits of transfer learning with a
  unified text-to-text transformer}.
\newblock \bibinfo{journal}{\emph{The Journal of Machine Learning Research}}
  \bibinfo{volume}{21}, \bibinfo{number}{1} (\bibinfo{year}{2020}),
  \bibinfo{pages}{5485--5551}.
\newblock


\bibitem[Sanh et~al\mbox{.}(2019)]%
        {sanh2019distilbert}
\bibfield{author}{\bibinfo{person}{Victor Sanh}, \bibinfo{person}{Lysandre
  Debut}, \bibinfo{person}{Julien Chaumond}, {and} \bibinfo{person}{Thomas
  Wolf}.} \bibinfo{year}{2019}\natexlab{}.
\newblock \showarticletitle{DistilBERT, a distilled version of BERT: smaller,
  faster, cheaper and lighter}.
\newblock \bibinfo{journal}{\emph{arXiv preprint arXiv:1910.01108}}
  (\bibinfo{year}{2019}).
\newblock


\bibitem[Schlichtkrull et~al\mbox{.}(2018)]%
        {schlichtkrull2018modeling}
\bibfield{author}{\bibinfo{person}{Michael Schlichtkrull},
  \bibinfo{person}{Thomas~N Kipf}, \bibinfo{person}{Peter Bloem},
  \bibinfo{person}{Rianne Van Den~Berg}, \bibinfo{person}{Ivan Titov}, {and}
  \bibinfo{person}{Max Welling}.} \bibinfo{year}{2018}\natexlab{}.
\newblock \showarticletitle{Modeling relational data with graph convolutional
  networks}. In \bibinfo{booktitle}{\emph{The Semantic Web: 15th International
  Conference, ESWC 2018, Heraklion, Crete, Greece, June 3--7, 2018, Proceedings
  15}}. Springer, \bibinfo{pages}{593--607}.
\newblock


\bibitem[Shi et~al\mbox{.}(2023)]%
        {shi2023oversampling}
\bibfield{author}{\bibinfo{person}{Shuhao Shi}, \bibinfo{person}{Kai Qiao},
  \bibinfo{person}{Jie Yang}, \bibinfo{person}{Baojie Song},
  \bibinfo{person}{Jian Chen}, {and} \bibinfo{person}{Bin Yan}.}
  \bibinfo{year}{2023}\natexlab{}.
\newblock \bibinfo{title}{Over-Sampling Strategy in Feature Space for Graphs
  based Class-imbalanced Bot Detection}.
\newblock
\newblock
\showeprint[arxiv]{2302.06900}~[cs.CV]


\bibitem[Tan et~al\mbox{.}(2023)]%
        {tan2023botpercent}
\bibfield{author}{\bibinfo{person}{Zhaoxuan Tan}, \bibinfo{person}{Shangbin
  Feng}, \bibinfo{person}{Melanie Sclar}, \bibinfo{person}{Herun Wan},
  \bibinfo{person}{Minnan Luo}, \bibinfo{person}{Yejin Choi}, {and}
  \bibinfo{person}{Yulia Tsvetkov}.} \bibinfo{year}{2023}\natexlab{}.
\newblock \showarticletitle{BotPercent: Estimating Twitter bot populations from
  groups to crowds}.
\newblock \bibinfo{journal}{\emph{arXiv preprint arXiv:2302.00381}}
  (\bibinfo{year}{2023}).
\newblock


\bibitem[Veli{\v{c}}kovi{\'c} et~al\mbox{.}(2017)]%
        {velivckovic2017graph}
\bibfield{author}{\bibinfo{person}{Petar Veli{\v{c}}kovi{\'c}},
  \bibinfo{person}{Guillem Cucurull}, \bibinfo{person}{Arantxa Casanova},
  \bibinfo{person}{Adriana Romero}, \bibinfo{person}{Pietro Lio}, {and}
  \bibinfo{person}{Yoshua Bengio}.} \bibinfo{year}{2017}\natexlab{}.
\newblock \showarticletitle{Graph attention networks}.
\newblock \bibinfo{journal}{\emph{arXiv preprint arXiv:1710.10903}}
  (\bibinfo{year}{2017}).
\newblock


\bibitem[Volkova et~al\mbox{.}(2015)]%
        {volkova2015inferring}
\bibfield{author}{\bibinfo{person}{Svitlana Volkova}, \bibinfo{person}{Yoram
  Bachrach}, \bibinfo{person}{Michael Armstrong}, {and} \bibinfo{person}{Vijay
  Sharma}.} \bibinfo{year}{2015}\natexlab{}.
\newblock \showarticletitle{Inferring latent user properties from texts
  published in social media}. In \bibinfo{booktitle}{\emph{Proceedings of the
  AAAI Conference on Artificial Intelligence}}, Vol.~\bibinfo{volume}{29}.
\newblock


\bibitem[Wei and Nguyen(2019)]%
        {wei2019twitter}
\bibfield{author}{\bibinfo{person}{Feng Wei} {and} \bibinfo{person}{Uyen~Trang
  Nguyen}.} \bibinfo{year}{2019}\natexlab{}.
\newblock \showarticletitle{Twitter bot detection using bidirectional long
  short-term memory neural networks and word embeddings}. In
  \bibinfo{booktitle}{\emph{2019 First IEEE International conference on trust,
  privacy and security in intelligent systems and applications (TPS-ISA)}}.
  IEEE, \bibinfo{pages}{101--109}.
\newblock


\bibitem[Wolf et~al\mbox{.}(2020)]%
        {wolf2020transformers}
\bibfield{author}{\bibinfo{person}{Thomas Wolf}, \bibinfo{person}{Lysandre
  Debut}, \bibinfo{person}{Victor Sanh}, \bibinfo{person}{Julien Chaumond},
  \bibinfo{person}{Clement Delangue}, \bibinfo{person}{Anthony Moi},
  \bibinfo{person}{Pierric Cistac}, \bibinfo{person}{Tim Rault},
  \bibinfo{person}{R{\'e}mi Louf}, \bibinfo{person}{Morgan Funtowicz},
  {et~al\mbox{.}}} \bibinfo{year}{2020}\natexlab{}.
\newblock \showarticletitle{Transformers: State-of-the-art natural language
  processing}. In \bibinfo{booktitle}{\emph{Proceedings of the 2020 conference
  on empirical methods in natural language processing: system demonstrations}}.
  \bibinfo{pages}{38--45}.
\newblock


\bibitem[Wu et~al\mbox{.}(2023)]%
        {wu2023bottrinet}
\bibfield{author}{\bibinfo{person}{Jun Wu}, \bibinfo{person}{Xuesong Ye}, {and}
  \bibinfo{person}{Yanyuet Man}.} \bibinfo{year}{2023}\natexlab{}.
\newblock \showarticletitle{Bottrinet: A unified and efficient embedding for
  social bots detection via metric learning}. In \bibinfo{booktitle}{\emph{2023
  11th International Symposium on Digital Forensics and Security (ISDFS)}}.
  IEEE, \bibinfo{pages}{1--6}.
\newblock


\bibitem[Yan et~al\mbox{.}(2020)]%
        {yan2020tinygnn}
\bibfield{author}{\bibinfo{person}{Bencheng Yan}, \bibinfo{person}{Chaokun
  Wang}, \bibinfo{person}{Gaoyang Guo}, {and} \bibinfo{person}{Yunkai Lou}.}
  \bibinfo{year}{2020}\natexlab{}.
\newblock \showarticletitle{Tinygnn: Learning efficient graph neural networks}.
  In \bibinfo{booktitle}{\emph{Proceedings of the 26th ACM SIGKDD International
  Conference on Knowledge Discovery \& Data Mining}}.
  \bibinfo{pages}{1848--1856}.
\newblock


\bibitem[Yang et~al\mbox{.}(2013)]%
        {yang2013empirical}
\bibfield{author}{\bibinfo{person}{Chao Yang}, \bibinfo{person}{Robert
  Harkreader}, {and} \bibinfo{person}{Guofei Gu}.}
  \bibinfo{year}{2013}\natexlab{}.
\newblock \showarticletitle{Empirical evaluation and new design for fighting
  evolving twitter spammers}.
\newblock \bibinfo{journal}{\emph{IEEE Transactions on Information Forensics
  and Security}} \bibinfo{volume}{8}, \bibinfo{number}{8}
  (\bibinfo{year}{2013}), \bibinfo{pages}{1280--1293}.
\newblock


\bibitem[Yang et~al\mbox{.}(2020)]%
        {yang2020scalable}
\bibfield{author}{\bibinfo{person}{Kai-Cheng Yang}, \bibinfo{person}{Onur
  Varol}, \bibinfo{person}{Pik-Mai Hui}, {and} \bibinfo{person}{Filippo
  Menczer}.} \bibinfo{year}{2020}\natexlab{}.
\newblock \showarticletitle{Scalable and generalizable social bot detection
  through data selection}. In \bibinfo{booktitle}{\emph{Proceedings of the AAAI
  conference on artificial intelligence}}, Vol.~\bibinfo{volume}{34}.
  \bibinfo{pages}{1096--1103}.
\newblock


\bibitem[Yang et~al\mbox{.}(2022)]%
        {yang2022rosgas}
\bibfield{author}{\bibinfo{person}{Yingguang Yang}, \bibinfo{person}{Renyu
  Yang}, \bibinfo{person}{Yangyang Li}, \bibinfo{person}{Kai Cui},
  \bibinfo{person}{Zhiqin Yang}, \bibinfo{person}{Yue Wang},
  \bibinfo{person}{Jie Xu}, {and} \bibinfo{person}{Haiyong Xie}.}
  \bibinfo{year}{2022}\natexlab{}.
\newblock \showarticletitle{RoSGAS: Adaptive Social Bot Detection with
  Reinforced Self-Supervised GNN Architecture Search}.
\newblock \bibinfo{journal}{\emph{ACM Transactions on the Web}}
  (\bibinfo{year}{2022}).
\newblock


\bibitem[Ye et~al\mbox{.}(2023)]%
        {ye2023hofa}
\bibfield{author}{\bibinfo{person}{Sen Ye}, \bibinfo{person}{Zhaoxuan Tan},
  \bibinfo{person}{Zhenyu Lei}, \bibinfo{person}{Ruijie He},
  \bibinfo{person}{Hongrui Wang}, \bibinfo{person}{Qinghua Zheng}, {and}
  \bibinfo{person}{Minnan Luo}.} \bibinfo{year}{2023}\natexlab{}.
\newblock \showarticletitle{HOFA: Twitter Bot Detection with Homophily-Oriented
  Augmentation and Frequency Adaptive Attention}.
\newblock \bibinfo{journal}{\emph{arXiv preprint arXiv:2306.12870}}
  (\bibinfo{year}{2023}).
\newblock


\bibitem[Zeng et~al\mbox{.}(2019)]%
        {zeng2019graphsaint}
\bibfield{author}{\bibinfo{person}{Hanqing Zeng}, \bibinfo{person}{Hongkuan
  Zhou}, \bibinfo{person}{Ajitesh Srivastava}, \bibinfo{person}{Rajgopal
  Kannan}, {and} \bibinfo{person}{Viktor Prasanna}.}
  \bibinfo{year}{2019}\natexlab{}.
\newblock \showarticletitle{Graphsaint: Graph sampling based inductive learning
  method}.
\newblock \bibinfo{journal}{\emph{arXiv preprint arXiv:1907.04931}}
  (\bibinfo{year}{2019}).
\newblock


\bibitem[Zhang et~al\mbox{.}(2021)]%
        {zhang2021graph}
\bibfield{author}{\bibinfo{person}{Shichang Zhang}, \bibinfo{person}{Yozen
  Liu}, \bibinfo{person}{Yizhou Sun}, {and} \bibinfo{person}{Neil Shah}.}
  \bibinfo{year}{2021}\natexlab{}.
\newblock \showarticletitle{Graph-less neural networks: Teaching old mlps new
  tricks via distillation}.
\newblock \bibinfo{journal}{\emph{arXiv preprint arXiv:2110.08727}}
  (\bibinfo{year}{2021}).
\newblock


\bibitem[Zhao et~al\mbox{.}(2022)]%
        {zhao2022learning}
\bibfield{author}{\bibinfo{person}{Jianan Zhao}, \bibinfo{person}{Meng Qu},
  \bibinfo{person}{Chaozhuo Li}, \bibinfo{person}{Hao Yan},
  \bibinfo{person}{Qian Liu}, \bibinfo{person}{Rui Li}, \bibinfo{person}{Xing
  Xie}, {and} \bibinfo{person}{Jian Tang}.} \bibinfo{year}{2022}\natexlab{}.
\newblock \showarticletitle{Learning on large-scale text-attributed graphs via
  variational inference}.
\newblock \bibinfo{journal}{\emph{arXiv preprint arXiv:2210.14709}}
  (\bibinfo{year}{2022}).
\newblock


\end{thebibliography}

\newpage

\appendix
\section{Algorithm for training \ourmethod{}} \label{app:alg}
\renewcommand{\algorithmicrequire}{\textbf{Input:}}
\renewcommand{\algorithmicensure}{\textbf{Output:}}
\begin{algorithm}[H]
    \caption{\ourmethod{} Optimization Algotithm}
    \label{alg:1}
    \begin{algorithmic}[1]
        \REQUIRE{Twitter bot detection dataset $S$}
        \ENSURE{Optimized LM parameters $\Theta_{\text{LM}}$, Optimized GNN/MLP parameters $\Theta_{\text{GNN/MLP}}$}   

        \STATE initialize $\Theta_{\text{LM}}$, $\Theta_{
        \text{GNN/MLP}}$
        \FOR{each user $i \in S$}
            \STATE preprocess user textual sequence $\boldsymbol{t}_i$
            \STATE obtain user embedding $\boldsymbol{z}_i \leftarrow$ Equation \eqref{eq:0}
            
            \STATE $Loss_{\text{LM finetuning}} \leftarrow$ Equation \eqref{eq:1}
            \STATE $\Theta_{\text{LM}} \leftarrow$ BackPropagate ($Loss_{\text{LM finetuning}}$)
            \WHILE{$\Theta_{\text{LM}}$, $\Theta_{\text{GNN/MLP}}$ have not converged}
                \STATE $Loss_{\text{GNN/MLP}} \leftarrow$ Equation \eqref{eq:2}
                \STATE $\Theta_{\text{GNN/MLP}} \leftarrow$ BackPropagate ($Loss_{\text{GNN/MLP}}$)
                \STATE Update soft label $\boldsymbol{\overline{y}}_i \leftarrow$ Equation \eqref{eq:3} 

                \STATE $Loss_{\text{LM}} \leftarrow$ Equation \eqref{eq:4}
                \STATE $\Theta_{\text{LM}} \leftarrow$ BackPropagate ($Loss_{\text{LM}}$)
                \STATE Update user embedding $\boldsymbol{z}_i \leftarrow$ Equation \eqref{eq:0}
            \ENDWHILE
        \ENDFOR

        \RETURN $\Theta_{\text{LM}}$, $\Theta_{\text{GNN/MLP}}$
        
    \end{algorithmic}

\end{algorithm}

\section{dataset statistics} \label{app:dataset} 
\begin{table}[H]
    \centering
    \caption{Statistics of the 4 selected datasets}
    \begin{adjustbox}{max width=1\linewidth}
    \begin{tabular}{c|c|c|c|c}
    \toprule[1.5pt]
    \textbf{Dataset} & \textbf{Cresci-2015} & \textbf{Cresci-2017}& \textbf{Midterm-2018}& \textbf{TwiBot-20}\\ 
    \midrule[0.75pt]
        \# User &  5,301 &  14,368 &  50,538  &  229,580\\
        \# Human &  1,950 &  3,474 &  8,092  &  5,237\\
        \# Bot &  3,351 &  10,894 &  42,446  &  6,589\\
        \# Tweet &  2,827,757 &  6,637,615 &  0  &  33,488,192\\
        Graph & \CheckmarkBold & \XSolidBrush & \XSolidBrush & \CheckmarkBold \\
        \bottomrule[1.5pt]
    \end{tabular}
    \end{adjustbox}
    \label{tab:datasets}
\end{table}

\section{baseline details} \label{app:baselines}
\begin{itemize}
    \item \textbf{SGBot} \cite{yang2020scalable} leverages user metadata and engineered features from tweets and adopts random forest for scalable and generalizable bot detection.
    \item \textbf{BotHunter} \cite{beskow2018bot} extracts features from user's metadata and exploit random forest as classifier.
    \item \textbf{Kudugunta \textit{et al.}} \cite{kudugunta2018deep} combines synthetic minority oversampling (SMOTE) with undersampling techniques. Then the best result is achieved by  AdaBoost Classifier.  
    \item \textbf{LOBO} \cite{echeverri2018lobo} extracts 19 features from each users' metadata and tweets and adopts random forest for classification.
    \item \textbf{BotRGCN} \cite{feng2021botrgcn} encodes the users' tweets and description with RoBERTa and extract numerical and categorical features and employs RGCN for classification.
    \item \textbf{RGT} \cite{feng2022heterogeneity} leverages graph transformers under each type of relationship and semantic attention network across all relationships to learn users' representation.
    \item \textbf{SimpleHGN} \cite{lv2021we} is an improvement over GAT, making it suitable for heterogeneous graphs with different types of edges. 
    \item \textbf{GLNN} \cite{zhang2021graph} distills the knowledge of GNN into MLP to achieve fast inference without graph structure.
    \item \textbf{BIC} \cite{lei2022bic} integrates text and graph modality with a text-graph interaction module, with additional functionality to detect advanced bots with a semantic consistency model.
    \item \textbf{BotBuster} \cite{ng2022botbuster} uses mixture-of-experts approach where each expert analyzes a specific type of information to facilitate cross-platform bot detection. 
\end{itemize}

\section{implementation details} \label{app:imp}
We use PyTorch \cite{paszke2019pytorch}, PyTorch Geometirc \cite{fey2019fast}, scikit-learn \cite{pedregosa2011scikit}, and Transformers \cite{wolf2020transformers} library to implement our proposed \ourmethod{}. To facilitate reproducibility, hyperparameters of \ourmethod{} are shown in Table \ref{tab:hyperparameter}. Our experiments are run on one Tesla V100 GPU with 32 GB memory. 

\begin{table}[H]
    \centering
    \caption{Hyperparameter settings of \ourmethod{}}
    \begin{tabular}{c|c}
    \toprule[1.5pt]
    \textbf{Hyperparameter} & \textbf{Value}\\ 
    \midrule[0.75pt]
        optimizer &  AdamW\\
        LM model & RoBERTa\\
        GNN model & RGCN\\
        learning rate for LM & $10^{-5}$\\
        learning rate for GNN & $5 \times 10^{-4}$\\
        dropout for LM & $0.1$\\
        dropout for GNN & $0.4$\\
        L2 regularization for LM $\lambda_1$ & $10^{-2}$\\
        L2 regularization for GNN $\lambda_2$ & $10^{-5}$\\
        number of layers for GNN $L$ & $2$\\
        hidden size for GNN & $128$\\
        temperature of knowledge distillation $T$ & 3\\
        weight on soft label loss $\alpha$ & $0.5$\\
        finetuning epochs for LM & $5$\\
        \bottomrule[1.5pt]
    \end{tabular}
    \label{tab:hyperparameter}
\end{table}

\section{limitations and future work}
We identify two limitations in \ourmethod{}:
\begin{itemize}
    \item Since LM contains a vast number of parameters, training \ourmethod{} is more computationally intensive compared to graph-based methods. Consequently, \ourmethod{} requires more time to train, which makes it more challenging to apply on large-scale datasets such as TwiBot-22 \cite{feng2022twibot}.
    \item Due to the limitation of sequence length in language models (LMs), such as the 512 maximum input length of RoBERTa \cite{liu2019roberta}, \ourmethod{} cannot take into account all the information of users, resulting in a limited proportion of tweets being included in the input textual sequence.
\end{itemize}

Moving forward, we plan to answer the following questions: how to extend \ourmethod{} to large-scale datasets, how to make more comprehensive use of user information under limited input sequence length, and how to extract textual information more effectively so that LM can learn more effective representation.
\end{document}